\documentclass[journal,onecolumn]{IEEEtran}

\ifCLASSINFOpdf
\else
   \usepackage[dvips]{graphicx}
\fi
\usepackage{url}

\makeatletter
\let\NAT@parse\undefined
\makeatother
\usepackage{hyperref}  
\usepackage{graphicx}
\usepackage{subfig}
\usepackage{ntheorem}
\usepackage{cite}
\usepackage{color}
\usepackage{amssymb}
\usepackage{amsmath}
\newtheorem{theorem}{Theorem}
\newtheorem{prop}[theorem]{Proposition} 
\newtheorem{claim}{Claim}
 \newtheorem*{proof}{Proof}

\global\long\def\vct#1{\boldsymbol{#1}}%

\global\long\def\R{\mathbb{R}}%
\global\long\def\E{\mathbb{E}}%

\global\long\def\va{\boldsymbol{a}}%
\global\long\def\vc{\boldsymbol{c}}%
\global\long\def\vg{\boldsymbol{g}}%

\global\long\def\vq{\boldsymbol{q}}%
\global\long\def\vu{\boldsymbol{u}}%
\global\long\def\vw{\boldsymbol{w}}%
\global\long\def\vx{\boldsymbol{x}}%
\global\long\def\vy{\boldsymbol{y}}%

\global\long\def\mA{\boldsymbol{A}}%
\global\long\def\mC{\boldsymbol{C}}%
\global\long\def\mU{\boldsymbol{U}}%

\global\long\def\mW{\boldsymbol{W}}%

\global\long\def\T{\intercal}%

\begin{document}

\title{Training Dynamics of Nonlinear Contrastive Learning Model in the High Dimensional Limit}

\author{Linghuan Meng, Chuang Wang$^\ast$
\thanks{This paper was submitted on March 5, 2024. This work was supported by the Pioneer Hundred Talents Program of CAS under Grant Y9S9MS08 and the National Natural Science Foundation of China under Grant U20A20223.}
\thanks{Linghuan Meng (menglinghuan2021@ia.ac.cn) and Chuang Wang (wangchuang@ia.ac.cn) are with MAIS, Institute of Automation of Chinese Academy of Sciences, Beijing, China. Chuang Wang is also with School of Artificial Intelligence, University of Chinese Academy of Sciences, Beijing, China. $\ast$ Corresponding author: Chuang Wang.}
}

\maketitle

\begin{abstract}
This letter presents a high-dimensional analysis of the training dynamics for a single-layer nonlinear contrastive learning model. The empirical distribution of the model weights converges to a deterministic measure governed by a McKean-Vlasov nonlinear partial differential equation (PDE). Under L2 regularization, this PDE reduces to a closed set of low-dimensional ordinary differential equations (ODEs), reflecting the evolution of the model performance  during the training process. We analyze the fixed point locations and their stability of the ODEs unveiling several interesting findings. First, only the hidden variable's second moment  affects feature learnability at the state with uninformative initialization. Second, higher moments influence the probability of feature selection  by controlling the attraction region, rather than affecting local stability. Finally, independent noises added in the data argumentation  degrade performance but negatively correlated noise can reduces the variance of gradient estimation yielding better performance. Despite of the simplicity of the analyzed model, it exhibits a rich phenomena of training dynamics, paving a way to understand more complex mechanism behind  practical large models.

\end{abstract}


\IEEEpeerreviewmaketitle

\section{Introduction}

\IEEEPARstart{C}ontrastive learning (CL)  is a promising self-supervised strategy to learn semantic representations in an unsupervised way. 
The model is trained by encouraging representations of different views from the same image to be similar and pushing representations of distinct images apart. This pre-trained model services as a fundamental backbone for downstream tasks, {\em e.g.,}  classification, segmentation and object recognition. 
Though it succeeded in numerous  applications, ranging from nature language processing \cite{devlin_bert_2019} to computer vision \cite{chen_simple_2020,he_momentum_2020,grill_bootstrap_2020,zbontar_barlow_2021,bardes_vicreg_2022} 
, theoretical understanding on how contrastive learning works is relatively limited. 

A set of existing  works aim to find common theoretical ground of variety of contrastive methods and to unify those models via information theory \cite{saunshi_theoretical_2019, bardes_vicreg_2022} or gradient expansion \cite{tao_exploring_2022}, yielding more variants of contrastive loss functions. 
Another line of works started from a toy but theoretically-manageable model to investigate fundamental mechanism  of contrastive learning, for example, the generalization ability \cite{haochen_provable_2021}, effect of data augmentation \cite{huang_towards_2023}, training convergence \cite{tian_understanding_2022} and mode collapsing \cite{jing_understanding_2022}. Those works  employed either  static analysis that studied the  landscape of loss function \cite{tian_understanding_2022-1} or  deterministic dynamic theory via the gradient flow. The latter modeled the stochastic training process of SGD by a deterministic continuous-time ODE in the small learning rate limit, in which all randomness in the training process are neglected and the analysis only reflects the scenario with the full-batch gradient and infinite-many training samples. 

In this work, we study the one-pass training dynamics of  the single-layer nonlinear CL model in the so-called mean-field limit or high-dimensional limit, where we keep the learning rate as a constant, and let the dimension of data go to infinity. In such limit, the training dynamics is characterized as a continuous-time stochastic process in the microscopic view characterizing the evolution of model weights, of which the probability distribution is governed by a McKean-Vlasov partial differential equation. By this way, we are able to study the effect of inaccurate estimation of gradient resulting from the mini-batch SGD and the finite learning rate, as well as the contribution of additive noises in the data argumentation operation. The major  contributions are as follows.

\begin{itemize}
\item We provide an exact PDE characterization on the weight evolution of the CL training dynamics  in the high-dimensional limit.

\item We study the trainability of a feature at random initial state and find it only depends on the second moments of the corresponding hidden variable, but the final selected feature depends on  higher moments as well.

\item We discover that independent noises added in both  branches  degrades the performance, but  if the noises are negatively correlated with proper strength, the gradient variance will be reduced, which improves the model. 
\end{itemize}

    The major technique used in this work, high-dimensional analysis, recently become prevailing in study online algorithmsin signal processing  \cite{wang2018subspace,wang2017scaling}  and training dynamics of many neural networks both for supervised \cite{goldt_dynamics_2019, mei_mean_2018} and unsupervised \cite{wang_solvable_2019,khemakhem_variational_2020} models. This technique can be dated back to dynamical mean-field theory in 1990s \cite{saad1995exact}. Its rigorous math  theory related to propagation of chaos \cite{keller_stochastic_1973}, which is  rigorously proved in the applications in signal processing and machine learning \cite{wang_scaling_2017}. Very recently,  this asymptotic modeling  \cite{veiga_phase_2022} is applied to different scaling settings. The scaling of constant learning rate, study in this work,  is at an edge of a phase transition between trivially easy and unrecoverable, where rich phenomena occurred  controlled not only by the learning rate, but also the batch size, noise level and signal strength of data models, etc. In this work, we will investigate the implications of these phenomena  for contrastive learning.

The remaining sections are organized as follows.  Section \ref{sec:model} describes the data  and CL model. In Section \ref{sec:dyn}, we introduce the main theory of PDE characterization for the training dynamics. In Section \ref{sec:ana}, we analyze the training dynamics by studying  stability of the fixed points of the differential equation. Finally, we conclude this work in Section \ref{sec:con}.

\section{  Models for Theoretical Analysis} \label{sec:model}
\subsection{Data Model}

We assume that a streaming of sample data $\vx_b\in \mathbb{R}^N$, $b=1,2,\ldots$,  are generated according to  the sphere model 
$
    \vx_b = \mA \left[ 
\begin{array}{c}
     \vc_b  \\
     \tilde{\va}_b 
\end{array}\right],
$
where $\mA \in \R^{N\times N}$ is an orthonormal matrix. The first $d_1$ columns of $\mA$ are considered as the feature vectors, and their corresponding hidden variables $\vc_b\in \mathbb{R}^{d_1}$  are drawn from a given distribution $P_{\vc}$. Background noise variables in the hidden space $\tilde{\va}_b\sim \mathcal{N}(0,I_{N-d_1})$ is drawn from the normal distribution.
This data model is equivalent to  
\begin{equation}
    \vx_b= \tfrac{1}{\sqrt{N}}\mU \vc_b+\va_b,
\end{equation}
where $\mU = \sqrt{N}\mA_{:,:d_1} \in \mathbb{R}^{N \times d_1}$, of which each column represents a feature vector. The hidden variable $\vc_b$ is the same as above, and $\va_b=\mA_{:,d_1:}\tilde{\va}_b $ is the background noise on the complementary space  with the distribution $ \mathcal{N}(0,I-\frac{1}{N}\mU \mU^\T)$. 

This data model is a toy but theoretically manageable model to describe complicated situations of the training dynamics of neural networks. Similar data models were used to study the dynamics of CL \cite{wen_toward_2021,yang_understanding_2023} and the static loss landscape \cite{saunshi_theoretical_2019}. 

\subsection{One-Layer Nonlinear Model for Contrastive Learning}
\begin{figure}[t]
\centerline{\includegraphics[width=\columnwidth]{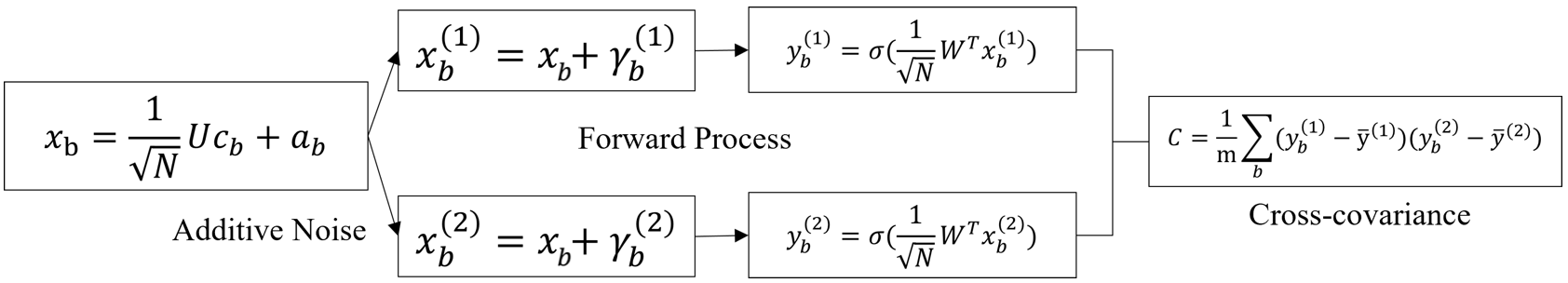}}
\caption{Structure of the 1-layer nonlinear contrastive learning model. For a batch of data, two additive noises are injected to each sample as two argumented views, which are  fed into the two branches of a one-layer nonlinear network respectively. 
}
\vspace{-1em}
\end{figure}

We introduce a 1-layer nonlinear model with the loss  similar to SimSiam \cite{chen_exploring_2021} to maximizing moments of activation and minimizing correlation  in the hidden spaces  . 

Given a sample $\vx_b$, we add two noises $ \vct{\gamma}^{(1)}, \vct{\gamma}^{(2)}$ in the two branches respectively as the data augmentation operation
\begin{equation*}
    \vx_b^{(1)} = \vx_b + \vct{\gamma}^{(1)}, \quad \quad \vx_b^{(2)} = \vx_b + \vct{\gamma}^{(2)}.
\end{equation*}

Then the two views of augmented data are fed into a single layer nonlinear neural network
\begin{equation*}
    \vy_b^{(1)} = \sigma(\tfrac{1}{\sqrt{N}}\mW^{\T}\vx_b^{(1)}),\quad
    \vy_b^{(2)} = \sigma(\tfrac{1}{\sqrt{N}}\mW^{\T}\vx_b^{(2)}),
\end{equation*}
where $\sigma$ is the activation function,  and  $\mW\in \mathbb{R}^{N \times d_2}$ represents the network weights.

The cross-covariance matrix of $\vy_b^{(1)}$, $\vy_b^{(2)}$ along the batch dimension is defined as

\begin{equation} \label{eq:C}
    \mC = \tfrac{1}{m}\textstyle{\sum}_{b=1}^m \big(\vy_b^{(1)}-\bar{\vy}^{(1)})(\vy_b^{(2)}-\bar{\vy}^{(2)} \big)^\T,
\end{equation}
where the mean  $\bar{\vy}_b^{(1)}$ is estimated practically by the batch average but we replace it by the population mean to avoid extra gradient path in the theoretical analysis.

The loss  is defined as
\begin{equation}\label{eq:loss}
    F(\mW;\{\vx_b\}) =  -\textstyle\sum_{i=1}^{d_2} C_{ii} + \lambda \textstyle\sum_{i\neq j} g( C_{ij})
\end{equation}
where $g:\mathbb{R} \to \mathbb{R}^+$ and $\lambda$ is a positive parameter  balancing between maximizing diagonal terms $C_{ii}$ and minimizing off-diagonal terms $C_{ij}$.

We study the training process using online stochastic gradient descent. At $k$th iteration step, the update rule is
\begin{equation}
    \begin{aligned}
    \tilde{\mW}_k &= \mW_k - \tau \tfrac{\partial F }{\partial \mW_k}-\tfrac{\tau}{N}\phi(\mW_k)
    \\
    \mW_{k+1,:,i} &= \tfrac{\sqrt{N}}{\Vert \tilde{\mW_k}_{:,i} \Vert_2}\tilde{\mW}_{k,:,i}  
    ,\quad \forall i = 1,2,\ldots,{d_2}
    \end{aligned}
\end{equation}
where   $\phi$ is an element-wise function as the gradient of  prior, and the second line is an additional column-wise normalization on the network weights for the convenient of analysis. Moreover, we use the one-pass (online) training paradigm. All data samples are used only once, where we generate a new batch of $m$ samples $\{\vx_b|b=1,2,\ldots,m\}$  for different step $k$.

We defined a minimal 1-layer nonlinear backbone model with  only  additive noises  used as data argumentation, but the theoretical analysis and the rich phenomena revealed in the subsequent sections shed light on understanding the training dynamics of more complex models, {\em e.g.} SimSiam,VicReg.

\section{ High-dimensional Training Dynamics} \label{sec:dyn}

We present an asymptotic  analysis of the training dynamics yielding a PDE characterization on the evolution of weight empirical distribution, and a low-dimensional  ODE for the macroscopic order parameters to reflect the model performance as a function of iteration number.  

\subsection{Simplifications and Assumptions } \label{sec:ass}

A.1  {Single-channel model for multiple data features:} We assume $d_2=1$ with a general  $d_1 \geq 1$.  The weight matrix $\mW_k$ then reduces to a vector $\vw_k \in \R^{N }$, and the last decorrelation term in the loss  \eqref{eq:loss} is omitted. Despite this simplification, we will eveal an interesting phenomenon that only one single feature rather than a mixture of multi-features can be learned.

A.2 {Population mean for batch normalization:} We replace the population mean instead of the batch mean to compute $\bar{y}$ in \eqref{eq:C}. This change simplifies the asymptotic equation without affecting qualitatively the  property of the dynamics. 

A.3 {Bounded moments:}  We assume that $\mU$ and $\vc_k$ are i.i.d. random variables with all  moments bounded.  

\subsection{Performance metric and Order Parameters}
We define the cosine similarity between the neural weight vector $\vw_k$ and feature matrix $\mU$  to measure the performance of the training model 
\begin{equation} \label{eq:cs}
    \vq^N_k \overset{\text{def}}{=} \tfrac{1}{N}\mU^\T \vw_k. 
\end{equation}
The $i$th entry of $\vq_k$ indicates how close the neural weight to $i$th feature ($i$th column of $\mU$) at $k$th  iteration. 

Moreover, we introduce another macroscopic quantity 
\begin{equation} \label{eq:rk}
    r^N_k \overset{\text{def}}{=} \tfrac{1}{N}\vw_k^\T \phi(\vw_k). 
\end{equation}

\subsection{PDE for joint distribution of feature and network weights}

To study the asymptotic training dynamics, we define the joint empirical measure  of  the model weight vector $\vw_k$ and the corresponding feature matrix $\mU $ as
\begin{equation*}
    \mu_{t}^N(w,\vu) \overset{\text{def}}{=} \tfrac{1}{N} \textstyle  \sum_{i=1}^{N} \delta(w - w_{k,i},\vu - \vu_{i}),
\end{equation*}
where $w_{k,i}$ and $ \vu_{i}$ are $i$th row of $\vw_k$ and $\mU$ respectively, and  $t \in \R^{+}$ is the continuous time index connecting the discrete step $k$ via the piece-wise constant interpolation $k = \lfloor tN \rfloor$.

The empirical distribution $\mu^N_t$ describes how weights evolves during the training. It also contains information on the model performance. For example, the cosine similarity \eqref{eq:cs} at iteration step $k$ can by expressed as an expectation w.r.t. this empirical measure, $\vq_t^N(t) =\E_{w,\vu \sim \mu^N_t} [w \vu] $ with $k = \lfloor tN \rfloor$. 

\begin{theorem} \label{thm:PDE}
Under the assumptions in Section \ref{sec:ass}, as $N \to \infty$, the sequence of random probability measures $\{\mu^N_t\}$ converges weakly to a deterministic measure $\mu_t$, of which the limiting density function $P_t(w,\vu)$ is the unique solution to the following nonlinear PDE    
\begin{equation}
    \partial_t P_t(w,\vu) = - \partial_{w} [\Gamma P_t(w,\vu)]
    +\tfrac{1}{2} \Lambda \partial^2_{w}P_t(w,\vu),
\end{equation}
where the drift and diffusion coefficients are 
\begin{equation}\label{eq:14}
\begin{aligned}
    \Gamma &= w[\vq^\T \vg+\tau r - \tfrac{1}{2}\Lambda] - \vu \vg-\tau\phi(w)
\\
     \Lambda &= 
         \tfrac{\tau^2}{4m} \Big[  \big(1+\langle (\gamma^{(2)})^2\rangle \big)\langle f_{12}^2\rangle +
         \big(1+\langle (\gamma^{(1)})^2\rangle\big) \langle f_{21}^2 \rangle
         \\
         &\quad \quad \quad+ 2\big(1+\langle \gamma^{(1)}\gamma^{(2)}\rangle \big) \langle f_{12} f_{21}\rangle    \Big],
    \end{aligned}
\end{equation}
respectively, and 
\begin{equation}\label{eq:qlim}
    \begin{aligned}
        \vq_t &= \textstyle \iint \vu^\T w P_t\text{d}w\text{d}\vu,
        \quad
        r_t = \textstyle \iint w \phi(w) P_t\text{d}w\text{d}\vu
    \end{aligned}
\end{equation}

with the symbols $\vg$, $f$, and $f^{\prime}$ are  shorthand for 
\begin{equation} \label{eq:short}
    \begin{aligned}
    \vg &= \tfrac{\tau}{2} \vq (\langle f^\prime_{12}\rangle+\langle f^\prime_{21}\rangle)  -  \tfrac{\tau}{2}\langle\vc f_{12}+\vc f_{21}\rangle  
     \\
     f_{\ell,\tilde{\ell}} &= -2 \big(\sigma(\Theta_\ell)-\bar{y}^{(\ell) } \big) \sigma^\prime (\Theta^{(\ell)} ),
     \quad \bar{y}^{(\ell)} = \langle \sigma( \Theta_{\ell} ) \rangle
     \\
     f^\prime_{\ell,\tilde{\ell}} &= \big(\tfrac{\partial}{\partial \Theta_{1}} + \tfrac{\partial}{\partial \Theta_{2}} \big) f_{\ell,\tilde{\ell}};\quad \ell=1,2
     \\
     \Theta_{\ell} &= \vc^\T\vq +  \sqrt{1-\vq^\T\vq} e + \vct{\gamma}^{(\ell)}, \;  e\sim\mathcal{N}(0,1)
    \end{aligned}
\end{equation}
The expect $\langle \cdot \rangle$ is taken w.r.t. $e\sim \mathcal{N} (0,1)$, the hidden variable $\vc\sim P_{\vc}$ and the additive noises $\gamma^{(\ell)}$. 
\end{theorem}

 The above PDE \eqref{eq:14} is a set of 1-D PDEs w.r.t. $w$  indexed by $\vu$. 

Given the order parameters $\vq_t$, $r_t$, $\bar{y}_t$, the set of PDEs are separated, in the sense that each PDE evolves independently. The joint probability are dependent only via the order parameters, which is an asymptotically independent high-dimensional process,  known as propagation of chaos \cite{keller_stochastic_1973}.  

Theorem \ref{thm:PDE} implies that the cosine similarity $\vq^N_k$ \eqref{eq:cs} and $r^N_k$ \eqref{eq:rk}  converges to the limits  in   \eqref{eq:qlim}, presenting a theoretical approach to tracking model performance during training.

\subsection{ODEs for the Cosine Similarity}
When no regularization other than L2 is imposed, $\phi(x)$ is linear,  and the limiting cosine similarity  $\vq_t$ forms a closed set of ordinary differential equations (ODEs). 
\begin{theorem}
    As $N \to \infty$, $q_t$ is the solution to the ODEs
    \begin{equation}
    \label{eq:16}
        \tfrac{\text{d}\vq}{\text{d}t} = \vq G - \vg,
    \end{equation}
    \label{theorem2}
\end{theorem}
with $G = \vq^\T \vg - \frac{1}{2}\Lambda$, and $\vg$ and $\Lambda$ defined in  \eqref{eq:14} and \eqref{eq:short}.

Theorem \ref{theorem2} implies that 
the macroscopic property of the training dynamics is described by the  ODEs when no regularization other than L2 
  are presented. In such cases, we can simplify the analysis by studying the local stability of the fixed points of the ODEs instead of  the PDE in Theorem~\ref{thm:PDE}.

\section{Analysis of the training dynamics}\label{sec:ana}
\label{Analysis}

We study how the hidden variable of feature and  additive noises affects  the training process. To further simplify the analysis, we remove the batch normalization by setting $\bar{y}=0$.

\subsection{Feature learnability depends on the second moment only}
\begin{figure}[t]

\subfloat[Training dynamics of $Q$]
 {
    \label{fig:subfig2}\includegraphics[width=0.47\columnwidth]{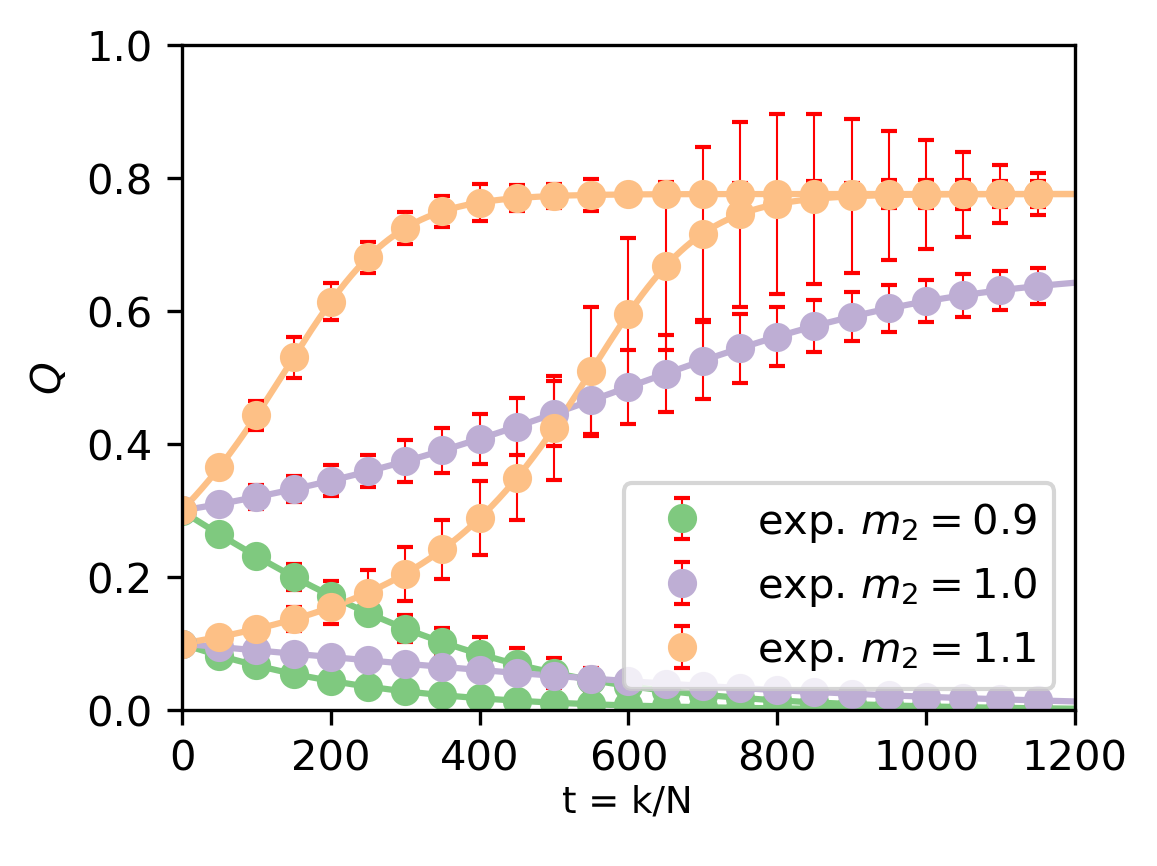}
 }\hfill
 \subfloat[r.h.s of the ODE v.s. $Q$]
 {
    \label{fig:subfig1}\includegraphics[width=0.47\columnwidth]{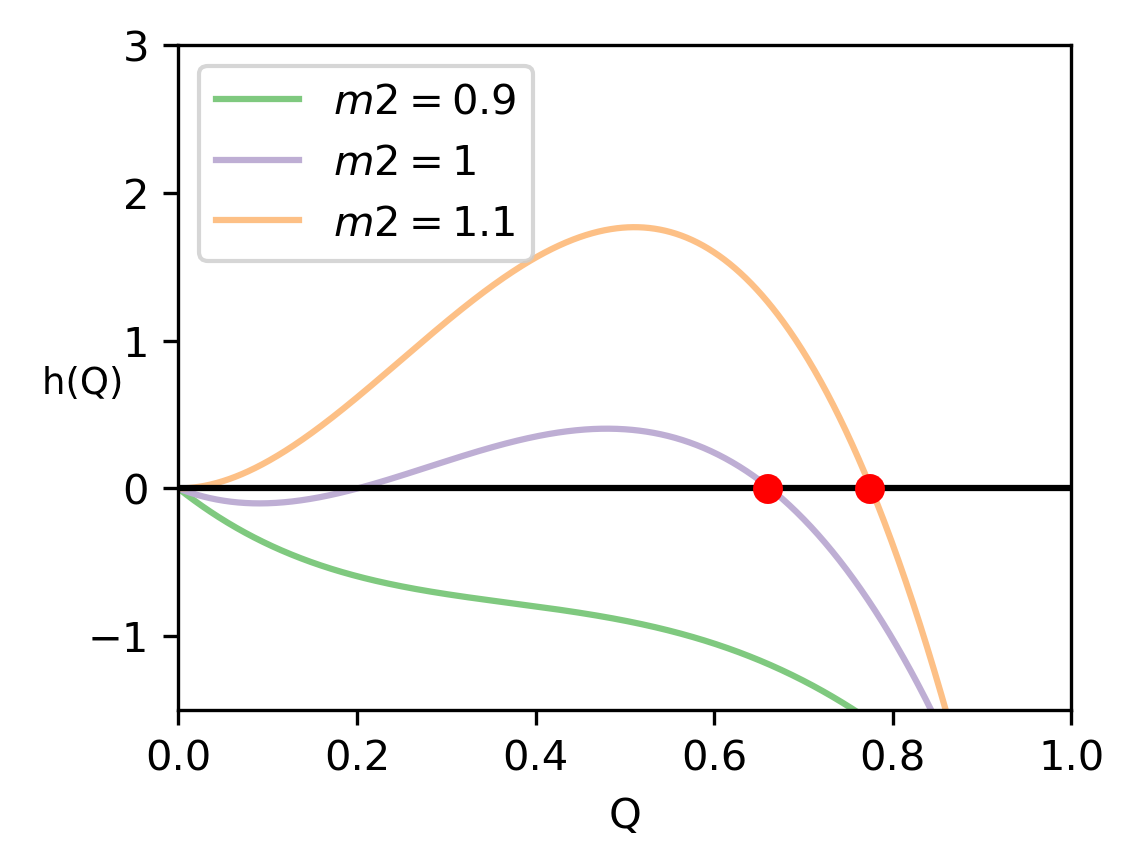}
 }

\caption{\label{fig:2}  (a) Dynamics of $Q_t$ with the theoretical prediction by \eqref{eq:ODE1d} (solid curves) and numerical simulations (errorbars)  (b) The r.h.s of the ODE   \eqref{eq:ODE1d} with $h(Q)=\frac{\text{d}Q}{\text{d}t}/\tau$. It visualizes the locations of fixed points (where cross the horizontal axis)  and their stability (stable if  derivative of $g(Q)$ is negative). 
}

\vspace{-2em}
\end{figure}

Whether  a feature can be  learned  relates to the local stability analysis of the ODE (\ref{eq:16}) at the initial point $q=0$ with $d_1=1$.

\begin{prop}
The fixed point of the ODE (\ref{eq:16}) at  $q=0$ is unstable if and only if
$$\langle c^2 \rangle 
> 1 + \frac{\tau}{2m}
 \frac{   (\langle \gamma^{(2)2} \rangle + 1) \langle\sigma_1^2\sigma_2'^2\rangle +(\langle \gamma^{(1)2} \rangle + 1) \langle \sigma_2^2\sigma_1'^2 \rangle +  2(1+\langle \gamma^{(1)} \gamma^{(2)} \rangle) \langle \sigma_1 \sigma_1' \sigma_2 \sigma_2'\rangle}{\langle \sigma_1 \sigma_2'' + 2\sigma_1'\sigma_2'+ \sigma_2 \sigma_1''  \rangle},$$
 with $\sigma_\ell = \sigma(e+\gamma^{(\ell)})$, and $\sigma^\prime_{(\ell)} = \sigma^\prime(e+\gamma^{(\ell)}), \ell=1,2$ and $\langle \cdot \rangle$ is the expect w.r.t. $e\sim \mathcal{N} (0,1)$ and the noises $\gamma^{(\ell)}$.
\end{prop}

This claim holds for any smooth activation function $\sigma(\cdot)$. A feature is learnable if the fixed point at $q=0$ is unstable. It implies that only the second moment $\langle c^2 \rangle $ of the hidden variable affects the learnability regardless to any choice of activation function, and higher moments do not affect the stability at the initialization state $q=0$. 

\paragraph*{Example}
    With the quadratic activation function $\sigma(x)=x^2$, $d_1=1$ and no additive noises, denoting $Q_t = q_t^2$, the ODE in Theorem \ref{theorem2} has an analytical expression
\begin{equation}\label{eq:ODE1d}
\begin{aligned}
    \tfrac{\text{d}Q}{\text{d}t} &= 
    8\tau (1-Q)(Q^2(m_4-3)+3Q(m_2-1)) 
    \\ & - 16 \tfrac{\tau^2}{m}Q(Q^3(m_6-15)+15Q^2(1-Q)(m_4-3)
    \\
    &+45Q(1-Q)^2 (m_2-1)+15) 
\end{aligned}
\end{equation}
where $m_2,m_4,m_6$ denotes the 2nd, 4th, 6th moments of the hidden variable $c$ respectively.

Fig.~\ref{fig:2} demonstrates the ODE theoretical prediction and numerical experiments with different second moments $m_2$ of $c$.  We generate the hidden value $c$  from the distribution
$
    P(c= \pm \sqrt{\alpha}) = b, P(c=0)=1-b,
$
where the moments are $m_2 = \alpha b,m_4 = \alpha^2 b, m_6 = \alpha^3 b$. The fixed point $Q=0$ is stable for $m_2=0.9$ or $1.0$ and  unstable for $m_2=1.1$.

\subsection{Higher moments affect feature selection probability as well}
We investigate the impact of higher moments on feature selection. 
 The analysis is restricted to a single-feature model $d_2=1$ with a quadratic activation function  $\sigma(x)=x^2$ on the data with  multiple features $d_1>1$. The limiting ODEs reveal that  only one feature can be learned, with the others completely discarded. Moreover,  higher moments influence the probability of which feature is selected.

We illustrate this phenomenon using a data model with two features.  The squared cosine similarity $Q_1(t)=q_1^2(t)$, $ Q_2(t)=q_1^2(t)$ forms a closed system of 2-D ODEs with an explicit form represented in the supplementary material (SM).
 Fig.~\ref{fig:4} visualizes training dynamics  as  the phase portrait, where the trajectories $(Q_1(t), Q_2(t))$ appear as non-crossing curves (blue curves).  In the experiment, the first feature is associated with a Gaussian hidden variable with  larger variable $m_2^{(1)}=1.2$, and second feature has a non-Gaussian hidden value with smaller variance $m_2^{(2)}=1.1$ but larger higher moments $m_4^{(2)}=6.05$, $m_6^{(2)} =33.28$.

Interestingly, the phase portrait in Fig.~\ref{fig:4} reveals that only one feature can be learned, as all stable fixed points (red and green dots) lie on the axes, and the fixed point (black dot) with non-vanishing correlation to both features is always unstable. Since the random initial state $Q_1=Q_2=0$ is an unstable fixed point as well, it is a random event to decide on which feature is learned  starting from this random uninformative initialization near the origin. Finally, the phase portrait shows that the attraction region to the second feature is  larger than the first one, indicating that higher moments affect the probability of feature selection via controlling attraction region to different recovery fix points. This observation implies that the feature with the highest second moment may not necessarily be the one learned, as demonstrated in Fig. \ref{fig:subfig42}.

\begin{figure}[t] 
\vspace{-1em}
\centering
\subfloat[Phase portrait]
  {
    \label{fig:subfig41}\includegraphics[width=0.43\columnwidth]{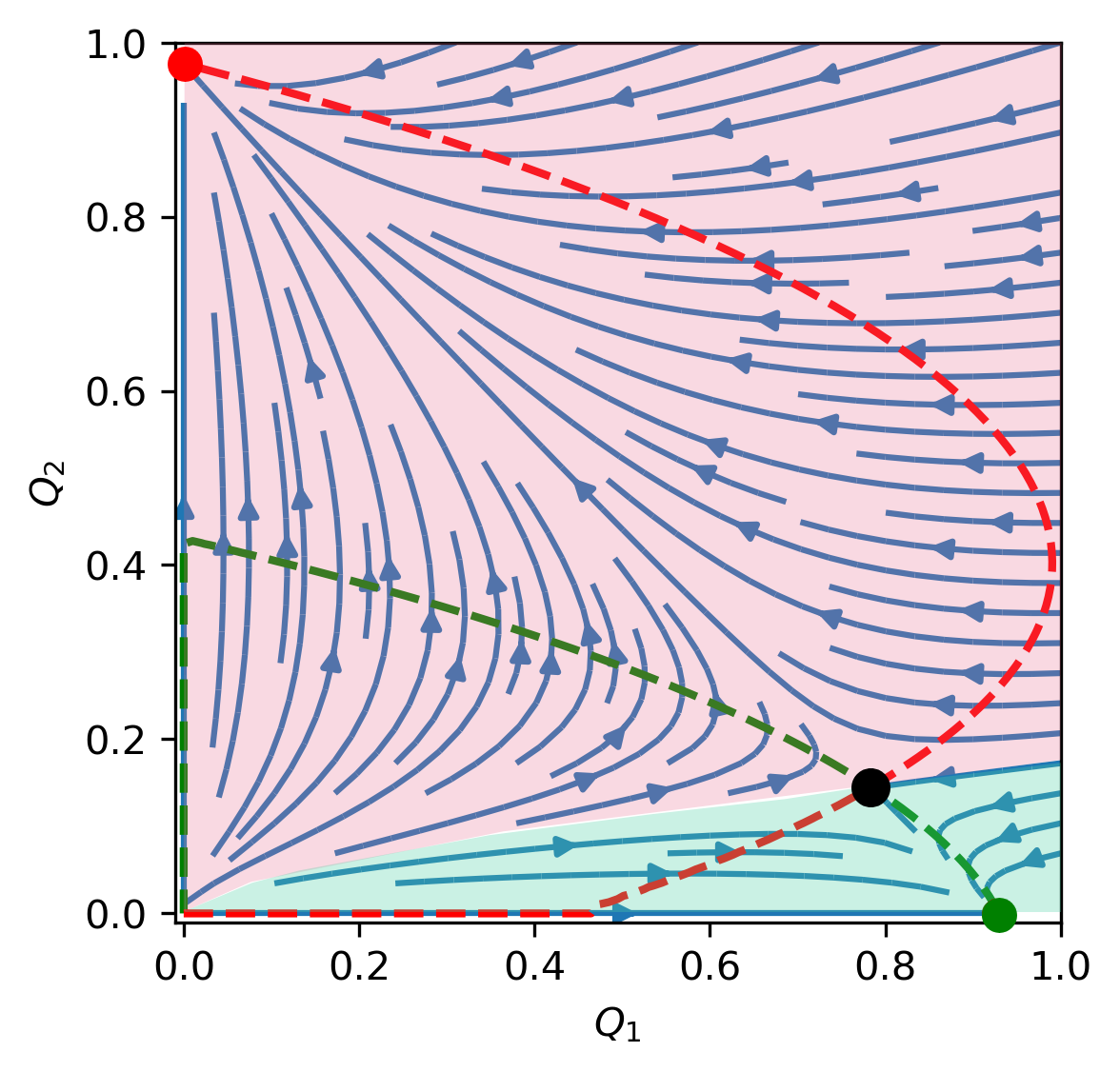}
  }
  \subfloat[Training dynamics]
  {
    \label{fig:subfig42}\includegraphics[width=0.57\columnwidth]{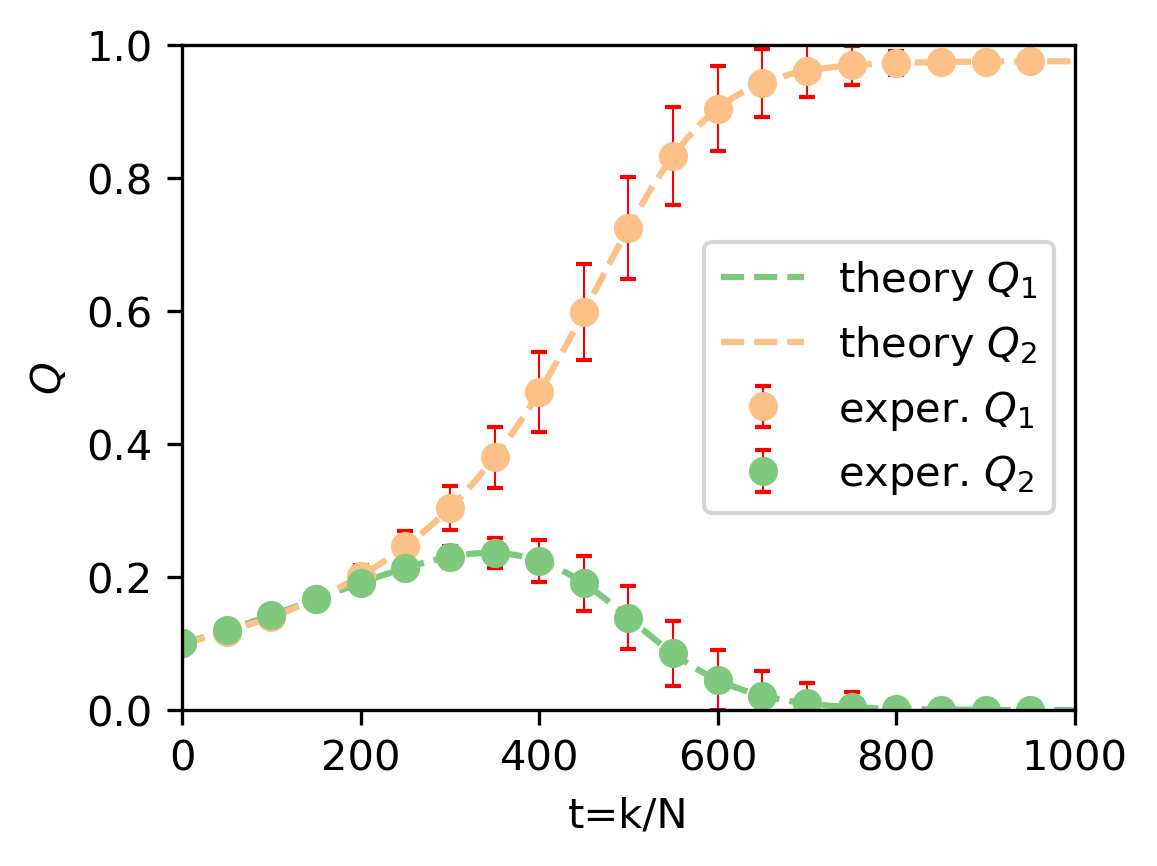}
  }\hfill
\caption{ (a) Phase portrait (b) Squared cosine similarities. The variance of hidden variables of the two features are $m_2^{(1)}=1.2$, $m_2^{(2)}=1.1$. Green- and red-shaddowed  areas are attraction regions to the first and second features respectively.}
\label{fig:4}
\vspace{-1em}
\end{figure}

\subsection{Correlated additive noise reduces gradient variance}

This section investigates the contribution of the additive noises to the impact  on the gradient variance and the subsequent effects on feature learning performance.

We still restrict the analysis to the quadratic activation function in order to get explicit analytical expressions, where the details are in SM. Our analysis reveals that imposing independent additive noises on the two branches consistently elevates the gradient variance compared to the noise-free scenario, resulting in a decline in feature recovery performance.

Conversely, we demonstrate the beneficial effects of correlated additive noises  both theoretically and numerically. The key factor appears to be the correlation between the injected noises, denoted by $\langle \gamma_1 \gamma_2\rangle$ in \eqref{eq:14}. If the two additive noises $\gamma_1$ and $\gamma_2$ are negatively correlated, the diffusion term $\Lambda$ in \eqref{eq:14}  can decrease.  Lowering the gradient variance with proper balance of noise correlation and strength, the training will facilitate a more precise estimation of the feature vector, leading to improved performance. 

\begin{figure}[t]
\centering
\includegraphics[width=0.65\columnwidth]{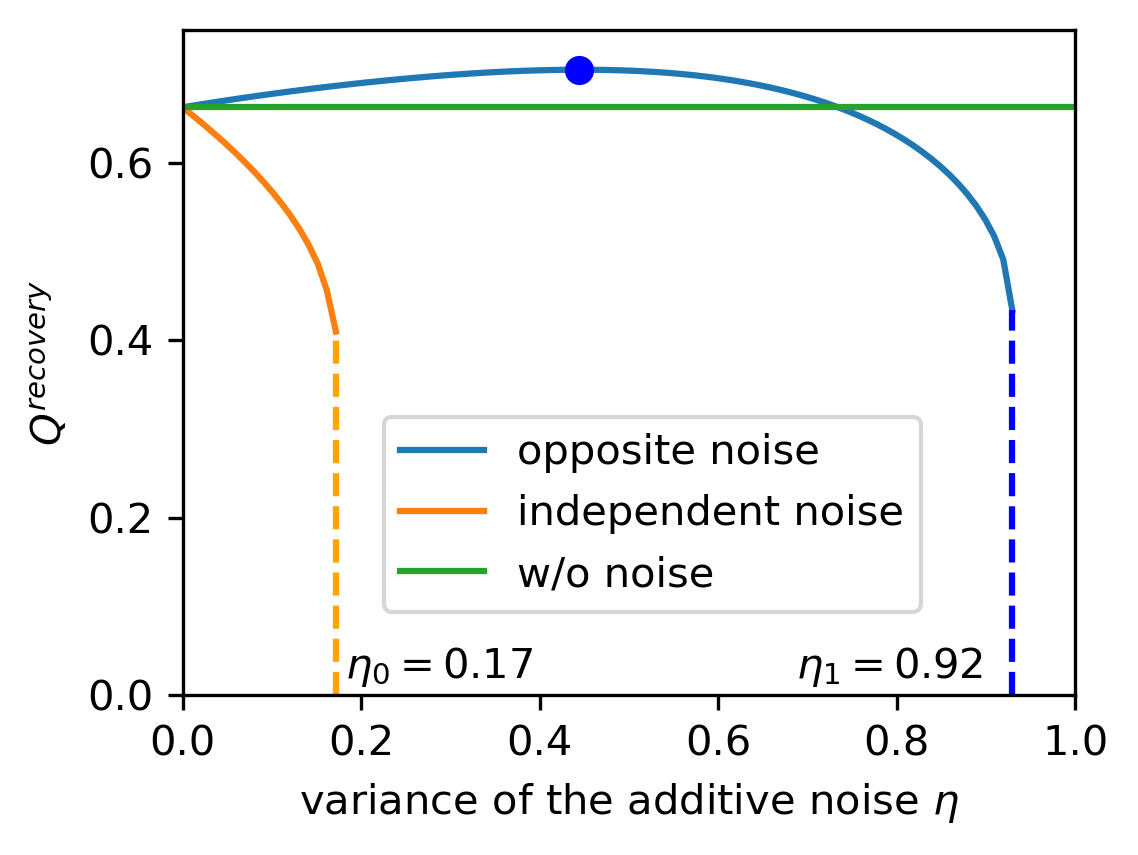}
\vspace{-1em}
\caption{ stationary cosine similarity v.s. strength additive noises at the recovery state }
\label{fig:6}
\vspace{-1.5em}
\end{figure}

Fig.~\ref{fig:6} illustrates the position of the fixed point corresponding to the feature recovery state (red dots in Fig.~\ref{fig:2}) as a function of the  variance  $\eta$ of the additive noise. 
Notably, while independent additive noise monotonically deteriorates performance, correlated noises injected into both branches enhance performance within a broad range of noise strength parameters $\eta$ (region where the blue curve is above the green line, with the optimal choice at blue point). 
Finally, the recovery state will vanish when $\eta$ reaches a critical threshold indicating a failure to learn. Similar phenomenon hold for other activation function, e.g. $\sigma(x)=\text{ReLu}(x)$.

\vspace{-1em}
\section{Conclusion}\label{sec:con}

This letter presents a high-dimensional analysis of the training dynamics for a  nonlinear contrastive learning model. 
The empirical distribution of model weights convergences to a deterministic measure governed by a McKean-Vlasov nonlinear PDE in the large system limit.  In the absence of additional regularization beyond L2, this PDE reduces to a closed set of ordinary differential equations (ODEs), reflecting the evolution of model performance during training. 

We reveal several insightful phenomena by studying stability and attraction region of fixed point of the ODEs.
First, only the second moment of the hidden
variable impacts the learnability. Second, higher moments regulate the feature selection probability by controlling attraction region. Finally, while independent noises  degrade performance, negatively correlated additive noise aids in reducing gradient variance.

While the model here is minimalistic, it showcases a richness of training dynamics phenomena. Future work will extend this framework to multi-channel models with stacked-layer structure, aiming to elucidate the mechanisms underlying  training dynamics of self-/un-supervised deep neural networks.

\bibliographystyle{IEEEtran}


\newpage
\appendices



\renewcommand{\theequation}{S-\arabic{equation}}
\renewcommand{\thesection}{S-\Roman{section}}

This supplementary material is organized as follows.  In Section \ref{appendix-sec:1}, we present the derivation of the PDE in Theorem 1. In Section \ref{appendix-sec:2}, we provide the  derivation of the ODE in Theorem 2. Then Section \ref{appendix-sec:3} contains the details of the theoretical  analysis based on the ODE. In particular, in \ref{appendix-subsec:31}, we prove  Proposition 3 and demonstrate the example of one feature case.
Next, we analyze the two feature case in \ref{appendix-subsec:32}, and study the effect of additive noises in \ref{appendix-subsec:33}. Finally, we present additional experiment results with ReLu activation function.

\section{Derivation of PDE in Theorem 1}
\label{appendix-sec:1}
For sake of completeness, we first restate the data model and model setting, which are an extended version of Section II in the main text. Afterwards we present a detailed derivation of the PDE in Theorem 1, which is determined by the first and second moments of the weight increment of a single training iteration.
We omit the lengthy rigorous proof, which requires careful control of the growth of higher moments.
\subsection{Data Model}
We assume that a streaming of sample data, $\vx_b\in \mathbb{R}^N, b=1,2,\ldots$,  are generated according to  the sphered model 
\begin{equation*}
    \vx_b = \mA \left[ 
\begin{array}{c}
     \vc_b  \\
     \tilde{\va}_b
\end{array}\right],
\end{equation*}
where $\mA \in \R^{N\times N}$ is an orthonormal matrix. The first $d_1$ columns of $\mA$ are considered as the feature vectors, of which the corresponding hidden variables $\vc_b\in \mathbb{R}^{d_1}$ are drawn from a given distribution $P_{\vc}$, and $\tilde{\va}_b\sim \mathcal{N}(0,I_{N-d_1})$ is drawn from the normal distribution representing background noises distributed on the complementary space to the feature space.

The above model is equivalent to  
\begin{equation}
    \vx_b= \tfrac{1}{\sqrt{N}}\mU \vc_b+\va_b,
\end{equation}
where $\mU = \sqrt{N}\mA_{:,:d_1} \in \mathbb{R}^{N \times d_1}$, of which each column represents a feature vector. The hidden variable $\vc_{b}$ is the same as above, and $\va_b=\mA_{:,d_1:}\tilde{\va}_b $ is the background noise with the distribution $ \mathcal{N}(0,I-\tfrac{1}{N}\mU \mU^\T)$. 

This data model is a toy but theoretically manageable model to describe complicated situations of the training dynamics of neural networks. The hidden variable $\vc_b$ can affect the frequency and significance of feature vector. For example, when $c\in\{0,\pm 1\}$ has sparse property, the corresponding feature vector may vanish in a large portion of data.

\subsection{One-Layer Nonlinear Model for Contrastive Learning}
\begin{figure}[htbp]
\centerline{\includegraphics[width = 0.7\linewidth]{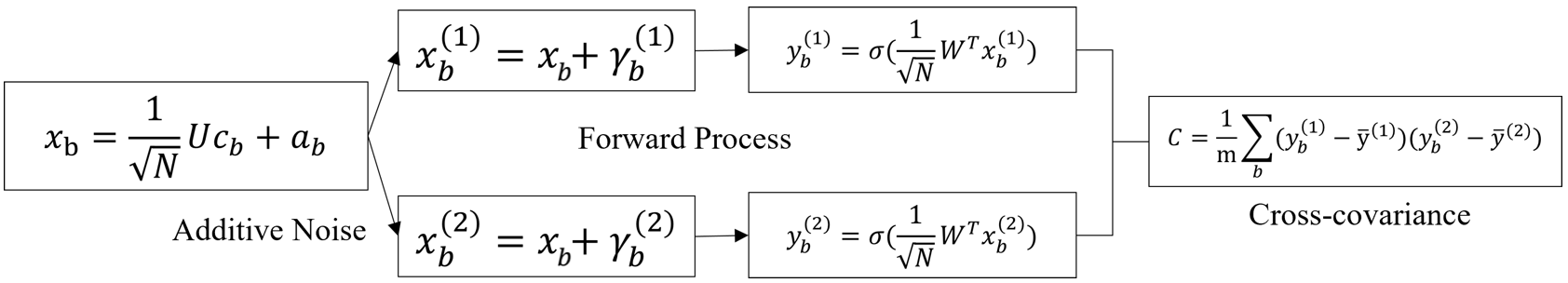}}
\caption{Structure of the 1-layer nonlinear contrastive model analyzed in this work.}
\label{appendix-fig:1}
\end{figure}
 The forward process is showed in Fig. \ref{appendix-fig:1}. For a stream of raw data $\vx_b\in \mathbb{R}^N,b=1,2,\ldots,m$ with batch size $m$, we first inject two noises  to the raw data $\vx_b$ respectively  as the input of  the two branches 
\begin{equation*}
    \vx_{b}^{(1)} = \vx_{b} + \vct{\gamma}^{(1)}, \quad \quad \vx_{b}^{(2)} = \vx_{b} + \vct{\gamma}^{(2)},
\end{equation*}
where $\vct{\gamma}^{(1)}$, $\vct{\gamma}^{(2)}$ are both N-D random vectors with zero mean representing additive noises imposed in the data argumentation procedure.

Then the two views of augmented data are fed into the the forward process of a single layer nonlinear neural network
\begin{equation*}
    \vy_{b}^{(1)} = \sigma(\tfrac{1}{\sqrt{N}}\mW^{\T}\vx_{b}^{(1)}),\quad
    \vy_{b}^{(2)} = \sigma(\tfrac{1}{\sqrt{N}}\mW^{\T}\vx_{b}^{(2)}),
\end{equation*}
where $\sigma$ is activation function, $\mW\in \mathbb{R}^{N \times d_2}$ is the model weight matrix, and $\vy_{b}^{(1)}\in \mathbb{R}^{d_2}$ is  the output activation.
The cross-covariance matrix between $\vy_{k}^{(1)}$, $\vy_{k}^{(2)}$ along the batch dimension is defined as 
\begin{equation*}
    \mC = \tfrac{1}{m}\textstyle{\sum}_{b=1}^m (\vy_{b}^{(1)}-\bar{\vy}^{(1)})(\vy_{b}^{(2)}-\bar{\vy}^{(2)})^\T.
\end{equation*}
Practically, the mean of $\vy_b^{(\ell)}$ is estimated by the batch average  $\bar{\vy}^{(\ell)}= \tfrac{1}{m}\sum_{b=1}^m \vy^{(\ell)}_b$ with $\ell=1,2$, but in the theoretical analysis, we replace it by the population mean to neglect extra gradient backward path introduced by this batch average.
In the theoretical analysis, we replace this mini-batch mean by the population mean to simplify the limiting PDE. 

The loss function  is defined as
\begin{equation*}
    F(X,w) =  -\textstyle\sum_{i=1}^{d_2} C_{ii} + \lambda \textstyle\sum_{i\neq j} g( C_{ij}),
\end{equation*}
where $\lambda$ is a positive parameter trading off the balance of diagonal terms $C_{ii}$ and off-diagonal terms $C_{ij}$ of the matrix $\mC$, and $g:\mathbb{R} \to \mathbb{R}^+$ is activation function, {\em e.g.} $g(x) = |x|$.

We study the training process using online mini-batch stochastic gradient descent
\begin{equation} \label{appendix-eq:itr}
    \begin{aligned}
    \tilde{\mW}_k &= \mW_k - \tau \tfrac{\partial F }{\partial \mW_k}-\tfrac{\tau}{N}\phi(\mW_k),
    \\
    \mW^{k+1}_{:,i} &= \tfrac{\sqrt{N}}{\Vert \tilde{\mW_k}_{:,i} \Vert_2}\tilde{\mW}_{k,:,i}  
    ,\quad \forall i = 1,2,\ldots,{d_2},
    \end{aligned}
\end{equation}
where   $\phi:\mathbb{R}^{N\times d_2} \to \mathbb{R}^{N\times d_2}$ is an element-wise function as the gradient of  prior, and the second line is an additional column-wise normalization on the network weights for the convenient of analysis.

We use the one-pass (online) training paradigm that all data samples are used only once, where we generate a batch of $m$ samples $\{\vx_b, b=1,2,\ldots,m\}$  for different step $k$.

\subsection{Explicit form of the gradients in SGD training process }
As the assumptions A.1 in Section III of the main, we will restrict the derivations to single-neural  model, {\em i.e.} $d_2=1$. In that case, the covariance matrix will degenerate into a scalar and the weight matrix $\mW_k$ will degenerate into a vector $\vw_k$

 Using the chain rule, we can get the explicit expression of the gradient
\begin{equation}
\begin{aligned}
    \tfrac{\partial F}{\partial w_{k,i}} 
    &=\tfrac{\partial F}{\partial C}\tfrac{\partial C}{\partial w_{k,i}}
    \\
    &= - \sum_{b=1}^m\tfrac{1}{m\sqrt{N}}((y_b^{(1)}-\bar{y}^{(1)})y'^{(2)}_{b}x_{b,i}^{(2)}+(y_{b}^{(2)}-\bar{y}^{(2)})y'^{(1)}_{b} x_{b,i}^{(1)}),
\end{aligned}
\end{equation}
where $y'^{(1)}_{b} =\sigma'(\tfrac{1}{\sqrt{N}}\vw_k^\T \vx^{(1)}_{b})$ and $w_{k,i}$ is $i$th entry of the model vector $\vw_k$.
The off-diagonal term appears only in the $d \geq 2$ situation, which is not considered here. Define a shorthand function
\begin{equation}
    \mu(z_1,z_2)  \overset{\text{def}}{=} - 2(\sigma(z_1)-\bar{y})\sigma'(z_2).
\end{equation}
The gradient can be rewriten as
\begin{equation}
    \tfrac{\partial F}{\partial w_{k,i}} 
    =\sum_{b=1}^m\tfrac{1}{2m\sqrt{N}}
    \bigg[x^{(2)}_{b,i}
        \mu(\tfrac{1}{\sqrt{N}}\vw_{k}^\T \vx^{(1)}_{b},\tfrac{1}{\sqrt{N}}\vw_{k}^\T \vx^{(2)}_{b} )
        + 
        x^{(1)}_{b,i} \mu(\tfrac{1}{\sqrt{N}}\vw_{k}^\T \vx^{(2)}_{b},\tfrac{1}{\sqrt{N}}\vw_{k}^\T \vx^{(1)}_{b}
        ) 
    \bigg].
\end{equation}

Finally, the update rule of the mini-batch stochastic gradient descent  is 
\begin{equation}\label{appendix-sec:itr}
    \begin{aligned}
    \tilde{\vw}_{k} &= \vw_{k} -\tfrac{\tau}{2m\sqrt{N}}\sum_{b=1}^m\bigg[\vx^{(2)}_{b}
        \mu(\tfrac{1}{\sqrt{N}}\vw_{k}^\T \vx^{(1)}_{b},\tfrac{1}{\sqrt{N}}\vw_{k}^\T \vx^{(2)}_{b} )
        + 
       \vx^{(1)}_{b} \mu(\tfrac{1}{\sqrt{N}}\vw_{k}^\T \vx^{(2)}_{b},\tfrac{1}{\sqrt{N}}\vw_{k}^\T \vx^{(1)}_{b}
        ) 
    \bigg] -\tfrac{\tau}{N}\phi(\vw_{k}),
    \\
    \vw_{k+1} &= \tfrac{\sqrt{N}}{\Vert \tilde{\vw}_k \Vert_2}\tilde{\vw}_{k}  
    ,\quad \forall i = 1,2,\ldots,{d_2},
    \end{aligned}
\end{equation}
where we normalize the network weights $\tilde{\vw}_k$ by column after each iteration for the convenience of analysis. Here we use $b$ as the index of sample in a batch of data and $k$ as the step of SGD iteration.


\subsection{Mean and Variance Calculation}

In this subsection, we  compute the mean and variance of the increment of the weights defined as
$$\vct{\Delta}_k = \vw_{k+1} -\vw_k.$$
 The first two moments $\E[\vct{\Delta}_k]$ and $\E [ \vct{\Delta}_k \vct{\Delta}_k^\T] $ are the essential variables to determine the partial differential equations representing the evolution of the probability density stated in Theorem 1.The operator $\E[\cdot]$, here and in this section, represents the conditional expectation  over the hidden variable $\vc$ and background noises $\va$ in the data model for the batch of samples at $k$th step $\{\vx_b|b=1,2,m\}$ as well as the additive noises $\vct{\gamma}^{(1)},\vct{\gamma}^{(2)}$, given the condition of the current model weights $\vw_k$ and data feature matrix $\mU$. 

The gradient-descent step in \eqref{appendix-sec:itr} is
\begin{equation}
    \tilde{\vct{\Delta}}_k = \tilde{\vw}_k - \vw_k = -\tfrac{\tau}{2m\sqrt{N}}\sum_{b=1}^m[\vx^{(2)}_{b}\mu(\tfrac{1}{\sqrt{N}}\vw_k^\T \vx^{(1)}_{b},\tfrac{1}{\sqrt{N}}\vw_k^\T \vx^{(2)}_{b})+ \vx^{(1)}_{b}\mu(\tfrac{1}{\sqrt{N}}\vw_k^\T \vx^{(2)}_{b},\tfrac{1}{\sqrt{N}}\vw_k^\T \vx^{(1)}_{b}) ] -\tfrac{\tau}{N}\phi(\vw_{k}).
\end{equation}
For the sake of simplicity, in the remaining of this subsection, we omit the subscript $k$ in $\vw_k$ and $\vq_k$, which are written as $\vw$ and $\vq$ respectively, and the $i$th entry of the vector $\tilde{\vct{\Delta}}_k $ is written as $\tilde{\Delta}_i$ instead of $[\tilde{\vct{\Delta}}_k]_i$.

In what follows, we will first calculate $\mathbb{E}[\tilde{\Delta}_{i}]$ and $\mathbb{E}[\tilde{\Delta}_{i}\tilde{\Delta}_{j}]$, which is weight increment at the gradient step in \eqref{appendix-sec:itr}. Then we will deal with the weight increment with the normalization step.

For a single sample $\vx_b$, consider $y^{(1)}_b$
\begin{equation}
    \begin{aligned}
        y^{(1)}_{b} &= \sigma(\tfrac{1}{\sqrt{N}}\vw_k^\T \vx^{(1)}_{k})
        =\sigma(\tfrac{1}{N} \vc_b^\T\mU^\T\vw_k+\tfrac{1}{\sqrt{N}}(\va_b+\vct{\gamma}^{(1)})^\T \vw_k)
        \\
        &=\sigma(\vc_b^\T\vq_k+\tfrac{1}{\sqrt{N}}(\va_b+\vct{\gamma}^{(1)})^\T \vw_k).
    \end{aligned}
\end{equation}
Both $\mu(\tfrac{1}{\sqrt{N}}\vw_k^\T \vx^{(1)}_{b},\tfrac{1}{\sqrt{N}}\vw_k^\T \vx^{(2)}_{b})$ and $\mu(\tfrac{1}{\sqrt{N}}\vw_k^\T \vx^{(2)}_{b},\tfrac{1}{\sqrt{N}}\vw_k^\T \vx^{(1)}_{b})$ can be regarded as functions of $\vq,\vw$ and random variables $\va,\vc,\gamma^{(1)},\gamma^{(2)}$. For the sake of simplicity, we omit the subscript $k$ for $\vw_k,\vq_k$ in the following derivation. The conditional expectation with respect to $\vc$ and $\va$ given $\vw$ and $\mU$ of the gradient $\mathbb{E}[\tilde{\Delta_{i}}]$ is
\begin{equation}
\label{appendix-eq:8}
\begin{aligned}
    \mathbb{E}[\tilde{\Delta}_{i}] &= -\tfrac{\tau}{2\sqrt{N}}\mathbb{E}
    [\mu(\vc^\T\vq+\tfrac{1}{\sqrt{N}}(\va+\vct{\gamma}^{(1)})^\T \vw,\vc^\T\vq+\tfrac{1}{\sqrt{N}}(\va+\vct{\gamma}^{(2)})^\T \vw)(\tfrac{1}{\sqrt{N}}\mU_{i,:} \vc+a_i+\vct{\gamma}^{(2)}_{i})]
    \\
    &-\tfrac{\tau}{2\sqrt{N}}\mathbb{E}
    [\mu(\vc^\T\vq+\tfrac{1}{\sqrt{N}}(\va+\vct{\gamma}^{(2)})^\T \vw,\vc^\T\vq+\tfrac{1}{\sqrt{N}}(\va+\vct{\gamma}^{(1)})^\T \vw)(\tfrac{1}{\sqrt{N}}\mU_{i,:} \vc+a_i+\vct{\gamma}^{(2)}_{i})]-\tfrac{\tau}{n}\phi(w_i)
    \\
    & = -\tfrac{\tau}{2\sqrt{N}}\mathbb{E}
    [\mu(\vc^\T\vq+\tfrac{1}{\sqrt{N}}(a_i+\gamma^{(1)}_{i}) w_i + \tilde{e}_{i} + \zeta^1_{i},\vc^\T\vq+\tfrac{1}{\sqrt{N}}(a_i+\gamma^{(2)}_{i}) w_i + \tilde{e}_{i} + \zeta^2_{i})(\tfrac{1}{\sqrt{N}}\mU_{i,:} \vc+a_i+\gamma^{(2)}_{i})]
    \\
    & -
    \tfrac{\tau}{2\sqrt{N}}\mathbb{E}
    [\mu(\vc^\T\vq+\tfrac{1}{\sqrt{N}}(a_i+\gamma^{(2)}_{i}) w_i + \tilde{e}_{i} + \zeta^2_{i},\vc^\T\vq+\tfrac{1}{\sqrt{N}}(a_i+\gamma^{(1)}_{i}) w_i + \tilde{e}_{i} + \zeta^1_{i})(\tfrac{1}{\sqrt{N}}\mU_{i,:} \vc+a_i+\gamma^{(1)}_{i})]
    \\&-\tfrac{\tau}{n}\phi(w_i),
\end{aligned}
\end{equation}
where $\tilde{e}_{i} \overset{\text{def}}{=} \tfrac{1}{\sqrt{N}} (\va^\T\vw-w_ia_i)$ and $\zeta^1_{i} \overset{\text{def}}{=} \tfrac{1}{\sqrt{N}} (\vct{\gamma}^{(1)\T}\vw-w_i\gamma^{(1)}_{i}),\zeta^2_{i} \overset{\text{def}}{=} \tfrac{1}{\sqrt{N}} (\vct{\gamma}^{(2)\T}\vw-w_i\gamma^{(2)}_{i})$. 

In high-dimensional setting, as $N\to \infty$, we use the Taylor expansion of $\mu$ around $\vc^\T\vq+ \tilde{e}_{i} + \zeta^2_{i}$ and $\vc^\T\vq+ \tilde{e}_{i} + \zeta^1_{i}$ up to the first order and compute the expectation term in (\ref{appendix-eq:8})
\begin{equation}
\label{appendix-eq:9}
    \begin{aligned}
        &\mathbb{E}
    [\mu(\vc^\T\vq+\tfrac{1}{\sqrt{N}}(\va+\gamma^{(1)})^\T \vw,\vc^\T\vq+\tfrac{1}{\sqrt{N}}(\va+\gamma^{(2)})^\T \vw)(\tfrac{1}{\sqrt{N}}\mU_{i,:} \vc+a_i+\gamma^{(2)}_{i})] \\
    & =\mathbb{E}
    [\mu(\vc^\T\vq+ \tilde{e}_{i} + \zeta^1_{i},\vc^\T\vq+ \tilde{e}_{i} + \zeta^2_{i})(\tfrac{1}{\sqrt{N}}\mU_{i,:} \vc+a_i+\gamma^{(2)}_{i})]
    \\
    &
    + \tfrac{1}{\sqrt{N}}w_i\mathbb{E}[
    \partial_1\mu(\vc^\T\vq+ \tilde{e}_{i} + \zeta^1_{i},\vc^\T\vq+ \tilde{e}_{i} + \zeta^2_{i})(\tfrac{1}{\sqrt{N}}\mU_{i,:} \vc+a_i + \gamma^{(1)}_{i})(a_i+\gamma^{(1)}_{i})]
    \\
    &+\tfrac{1}{\sqrt{N}}w_i\mathbb{E}[\partial_2\mu(\vc^\T\vq+ \tilde{e}_{i} + \zeta^1_{i},\vc^\T\vq+ \tilde{e}_{i} + \zeta^2_{i})(\tfrac{1}{\sqrt{N}}\mU_{i,:} \vc+a_i + \gamma^{(2)}_{i})(a_i+\gamma^{(2)}_{i})] + o(\tfrac{1}{\sqrt{N}}),
    \\
    \end{aligned}
\end{equation}
where $o(\tfrac{1}{\sqrt{N}})$ denotes all higher order terms. 
As $n\to \infty$, the random variable $\vq$ become deterministic. Moreover, random variables vector $\left[ \begin{array}{cc}
     \tilde{e}_{i}  &
      a_i   
\end{array}\right]^\T$ are all zero-mean Gaussian with the covariance matrix 
\begin{equation}
    \Sigma_{\tilde{e}} = \left[
    \begin{array}{cc}
        1-\vq^\T \vq +O(\tfrac{1}{N})& -\tfrac{1}{\sqrt{N}}\mU_{i,:}\vq   \\
        -\tfrac{1}{\sqrt{N}}\mU_{i,:}\vq  &  1+O(\tfrac{1}{N})  \\
    \end{array}
    \right].
\end{equation}

We assume all channels of $\gamma^{(1)}$  and that of $\gamma^{(2)}$ are i.i.d. random variables. Thus, the covariance matrix of $\left[ \begin{array}{cccc} 
     \tilde{\zeta}^1_{i}  &
      \tilde{\zeta}^2_{i} &
      \gamma^{(1)}_i&
      \gamma^{(2)}_i
\end{array}\right]^\T$ has the form
\begin{equation}
    \Sigma_{\gamma} = \left[
    \begin{array}{cccc}
        \langle (\gamma^{(1)})^2 \rangle&\langle \gamma^{(1)}\gamma^{(2)} \rangle&0&0 \\
        \langle \gamma^{(1)}\gamma^{(2)} \rangle&\langle (\gamma^{(1)})^2 \rangle&0&0 \\
        0&0&\langle (\gamma^{(1)})^2 \rangle&\langle \gamma^{(1)}\gamma^{(2)} \rangle \\
        0&0&\langle \gamma^{(1)}\gamma^{(2)} \rangle&\langle (\gamma^{(1)})^2 \rangle \\
    \end{array}
    \right],
\end{equation}
where $\langle \cdot \rangle$ means taking expectation over random variables $\vct{\gamma}^{(1)},\vct{\gamma}^{(2)},\va,\vc$. For multidimensional Gaussian distributions where the covariance matrix $\Sigma_{\tilde{e}}$ is not a unit matrix, we can orthogonalize the covariance matrix, thereby transforming the otherwise non-independent Gaussian into an independent Gaussian via a linear transformation.

We can introduced an independent random variables $e_{i} \sim \mathcal{N}(0,1)$, then we get
\begin{equation}
        \tilde{e}_{i}
        = \sqrt{1-\vq^\T\vq-(\tfrac{1}{\sqrt{N}}\mU_{i,:}\vq)^2}e_{i} - \tfrac{1}{\sqrt{N}}\mU_{i,:}\vq a_i
        = \sqrt{1-\vq^\T\vq}e_{i} -  \tfrac{1}{\sqrt{N}}\mU_{i,:}\vq a_i + o(\tfrac{1}{\sqrt{N}}).
\end{equation}
Thus using the Taylor expansion around $\vc^\T\vq+ \sqrt{1-\vq^\T\vq}e_{i} + \zeta^1_{i}$, the terms of (\ref{appendix-eq:9}) are simplified to

\begin{equation*}
    \begin{aligned}
        \mathbb{E}&
    [\mu(\vc^\T\vq+ \tilde{e}_{i} + \zeta^1_{i},\vc^\T\vq+ \tilde{e}_{i} + \zeta^2_{i})(\tfrac{1}{\sqrt{N}}\mU_{i,:} \vc+a_i+\gamma^{(2)}_{i})]\\
    &=\mathbb{E}
    [\mu(\vc^\T\vq+ \sqrt{1-\vq^\T\vq}e_{i} + \zeta^1_{i},\vc^\T\vq+ \sqrt{1-\vq^\T\vq}e_{i} + \zeta^2_{i})(\tfrac{1}{\sqrt{N}}\mU_{i,:} \vc+a_i+\gamma^{(2)}_{i})]
    \\
    &= \tfrac{1}{\sqrt{N}}\mU_{i,:} [\langle \vc \mu(\vc^\T\vq+ \sqrt{1-\vq^\T\vq}e_{i} + \zeta^1_{i},\vc^\T\vq+ \sqrt{1-\vq^\T\vq}e_{i} + \zeta^2_{i}) \rangle 
    \\
    &- \vq\langle \mu'(\vc^\T\vq+ \sqrt{1-\vq^\T\vq}e_{i} + \zeta^1_{i},\vc^\T\vq+ \sqrt{1-\vq^\T\vq}e_{i} + \zeta^2_{i}) \rangle] + O(\tfrac{1}{N});
    \\
    &
    \tfrac{1}{\sqrt{N}}w_i\mathbb{E}[
    \partial_1\mu(\vc^\T\vq+ \tilde{e}_{i} + \zeta^1_{i},\vc^\T\vq+ \tilde{e}_{i} + \zeta^2_{i})(\tfrac{1}{\sqrt{N}}\mU_{i,:} \vc+a_i + \gamma^{(2)}_{i})(a_i+\gamma^{(1)}_{i})]
    \\
    &=\tfrac{1}{\sqrt{N}}(1+\langle \gamma^{(1)}\gamma^{(2)}\rangle)w_i\langle \partial_1\mu(\vc^\T\vq+ \tilde{e}_{i} + \zeta^1_{i},\vc^\T\vq+ \tilde{e}_{i} + \zeta^2_{i})\rangle+ O(\tfrac{1}{N});
    \\
    &\tfrac{1}{\sqrt{N}}w_i\mathbb{E}[\partial_2\mu(\vc^\T\vq+ \tilde{e}_{i} + \zeta^1_{i},\vc^\T\vq+ \tilde{e}_{i} + \zeta^2_{i})(\tfrac{1}{\sqrt{N}}\mU_{i,:} \vc+a_i + \gamma^{(2)}_{i})(a_i+\gamma^{(2)}_{i})] 
    \\
    &=\tfrac{1}{\sqrt{N}}(1+\langle (\gamma^{(2)})^2\rangle)w_i\langle \partial_2\mu(\vc^\T\vq+ \tilde{e}_{i} + \zeta^1_{i},\vc^\T\vq+ \tilde{e}_{i} + \zeta^2_{i})\rangle+ o(\tfrac{1}{\sqrt{N}}).
    \end{aligned}
\end{equation*}
Noting that we have calculated the expectation over $\gamma^{(1)}_i,\gamma^{(2)}_i$, and the statistical properties of $\zeta^1_i,\zeta^2_i$ and $\gamma^{(1)}_i,\gamma^{(2)}_i$ are the same, thus we can substitute $\zeta^1_i,\zeta^2_i$ to $\gamma^{(1)},\gamma^{(2)}$. For the simplicity of the expression, we define the following symbols
\begin{equation}
\label{appendix-eq:14}
    \begin{aligned}
    \vg_i &= \tfrac{\tau}{2} \vq (\langle f^\prime_{12,i}\rangle+\langle f^\prime_{21,i}\rangle)  -  \tfrac{\tau}{2}\langle\vc f_{12,i}+\vc f_{21,i}\rangle  
     \\
     f_{\ell\tilde{\ell},i} &= -2 \big(\sigma(\Theta_{\ell,i})-\bar{y}^{(\ell)}_i \big) \sigma^\prime (\Theta_{\tilde{\ell},i} ),
     \quad \bar{y}^{(\ell)} = \langle \sigma( \Theta_{\ell,i} ) \rangle
     \\
     f^\prime_{\ell \tilde{\ell},i} &= \big(\tfrac{\partial}{\partial \Theta_{1}} + \tfrac{\partial}{\partial \Theta_{2}} \big) f_{\ell\tilde{\ell},i} 
     \\
     \partial_1 \mu_{\ell\tilde{\ell},i}&=\partial_1\mu(\Theta_{\ell,i},\Theta_{\tilde{\ell},i})
     \\
     \Theta_{\ell,i} &= \vc^\T\vq + e \sqrt{1-\vq^\T\vq}  + \vct{\gamma}_i^{(\ell)}, \; \ell=1,2.
    \end{aligned}
\end{equation}

With the abbreviation above,  (\ref{appendix-eq:9}) can be written as

\begin{equation}
    \begin{aligned}
    \mathbb{E}&
    [\mu(\vc^\T\vq+\tfrac{1}{\sqrt{N}}(\va+\gamma^{(1)})^\T \vw,\vc^\T\vq+\tfrac{1}{\sqrt{N}}(\va+\gamma^{(2)})^\T \vw)(\tfrac{1}{\sqrt{N}}\mU_{i,:} \vc+a_i+\gamma^{(2)}_{i})]
    \\
    &= -\tfrac{1}{\sqrt{N}} \mU_{i,:}[\langle \vc f_{12,i} \rangle - \vq\langle f'_{12,i}\rangle] + \tfrac{1}{\sqrt{N}}w_i[(1+\langle \gamma^{(1)} \gamma^{(2)}\rangle) \partial_1 \mu_{12,i}+(1+\langle (\gamma^{(2)})^2 \rangle) \partial_2 \mu_{12,i}].
    \end{aligned}
\end{equation}
Therefore, (\ref{appendix-eq:8}) is
\begin{equation}
    \begin{aligned}
        \mathbb{E}[\tilde{\Delta}_{i}] &= -\tfrac{\tau}{2\sqrt{N}}\mathbb{E}
    [\mu(\vc^\T\vq+\tfrac{1}{\sqrt{N}}(\va+\gamma^{(1)})^\T \vw,\vc^\T\vq+\tfrac{1}{\sqrt{N}}(\va+\gamma^{(2)})^\T \vw)(\tfrac{1}{\sqrt{N}}\mU_{i,:} \vc+a_i+\gamma^{(2)}_{i})]
    \\
    &-\tfrac{\tau}{2\sqrt{N}}\mathbb{E}
    [\mu(\vc^\T\vq+\tfrac{1}{\sqrt{N}}(\va+\gamma^{(2)})^\T \vw,\vc^\T\vq+\tfrac{1}{\sqrt{N}}(\va+\gamma^{(1)})^\T \vw)(\tfrac{1}{\sqrt{N}}\mU_{i,:} \vc+a_i+\gamma^{(2)}_{i})]-\tfrac{\tau}{N}\phi(w_i)
    \\
    & = \tfrac{1}{N}[-\mU_{i,:}\vg_i + \tfrac{\tau}{2} w_i[(1+\langle \gamma^{(1)}\gamma^{(2)} \rangle) \partial_1 \mu_{12,i}+(1+\langle (\gamma^{(2)})^2 \rangle) \partial_2 \mu_{12,i} + (1+\langle \gamma^{(1)}\gamma^{(2)} \rangle) \partial_1 \mu_{21,i}+(1+\langle (\gamma^{(1)})^2 \rangle) \partial_2 \mu_{21,i}]]
    \\
    &-\tfrac{\tau}{N} \phi(w_i) + o(\tfrac{1}{N}).
    \end{aligned}
\end{equation}

To compute the variance, noting that batch size affects the expected value of the variance, we get

\begin{equation*}
\begin{aligned}
    \mathbb{E}[\tilde{\Delta}_{i}^2] & =\mathbb{E}(
     -\tfrac{\tau}{m\sqrt{N}}\sum_{b=1}^m[x^{(2)}_{b,i}\tfrac{1}{2}\mu(\tfrac{1}{\sqrt{N}}\vw^\T \vx^{(1)}_{b},\tfrac{1}{\sqrt{N}}\vw^\T \vx^{(2)}_{b})+ x^{(1)}_{b,i}\tfrac{1}{2}\mu(\tfrac{1}{\sqrt{N}}\vw^\T \vx^{(2)}_{b},\tfrac{1}{\sqrt{N}}\vw^\T x^1_{b}) ] -\tfrac{\tau}{N}\phi(\vw_{k}))^2
     \\
     &= 
     \tfrac{\tau^2}{m^2N}\mathbb{E}(\sum_{b=1}^m[x^{(2)}_{b,i}\tfrac{1}{2}\mu(\tfrac{1}{\sqrt{N}}\vw^\T \vx^{(1)}_{b},\tfrac{1}{\sqrt{N}}\vw^\T \vx^{(2)}_{b})+ x^{(1)}_{b,i}\tfrac{1}{2}\mu(\tfrac{1}{\sqrt{N}}\vw^\T \vx^{(2)}_{b},\tfrac{1}{\sqrt{N}}\vw^\T x^1_{b}) ])^2 + o(\tfrac{1}{N})
    \\
     &= 
     \tfrac{\tau^2}{4m N}\mathbb{E}[(x^{(2)}_{i})^2\mu(\tfrac{1}{\sqrt{N}}\vw^\T \vx^{(1)}_{},\tfrac{1}{\sqrt{N}}\vw^\T \vx^{(2)}_{})^2+ (x^{(1)}_{i})^2\mu(\tfrac{1}{\sqrt{N}}\vw^\T \vx^{(2)}_{},\tfrac{1}{\sqrt{N}}\vw^\T x^1_{}))^2  
     \\
     &+ 2x^{(1)}_{i} x^{(2)}_{i}\mu(\tfrac{1}{\sqrt{N}}\vw^\T \vx^{(1)}_{},\tfrac{1}{\sqrt{N}}\vw^\T \vx^{(2)}_{}) \mu(\tfrac{1}{\sqrt{N}}\vw^\T \vx^{(2)}_{},\tfrac{1}{\sqrt{N}}\vw^\T \vx^{(1)}_{}) ] + o(\tfrac{1}{N})
    \\&=\tfrac{\tau^2}{4Nm} [(1+\langle (\gamma^{(2)})^2\rangle)\langle f^2_{12,i} \rangle+(1+\langle (\gamma^{(1)})^2\rangle)\langle f^2_{21,i} \rangle+(2+2\langle \gamma^{(1)} \gamma^{(2)}\rangle)\langle f_{12,i}f_{21,i} \rangle] + o(\tfrac{1}{N}).
\end{aligned}
\end{equation*}

If the index is different $i \neq j$, the expectation can be calculated separately, thus the covariance term is small quantity with a higher order
\begin{equation*}
    \begin{aligned}
        \mathbb{E}[\tilde{\Delta}_{i}\Tilde{\Delta}_{j}] &= \mathbb{E}[\Tilde{\Delta}_{i}]\mathbb{E}[\Tilde{\Delta}_{j}] = o(\tfrac{1}{N}).
    \end{aligned}
\end{equation*}

In summary, the covariance term is

\begin{equation}
\label{appendix-eq:17}
    \mathbb{E}[\tilde{\Delta}_{i}\Tilde{\Delta}_{j}] = \left\{ \begin{array}{cc}
         o(\tfrac{1}{N})& i\neq j  \\
         \tfrac{\tau^2}{4Nm} [(1+\langle (\gamma^{(2)})^2\rangle)\langle f^2_{12,i} \rangle+(1+\langle (\gamma^{(1)})^2\rangle)\langle f^2_{21,i} \rangle+(2+2\langle \gamma^{(1)} \gamma^{(2)}\rangle)\langle f_{12,i}f_{21,i} \rangle]+o(\tfrac{1}{N})&  i = j
    \end{array}\right.
\end{equation}

Next, we deal with the normalization step. Also, we use the Taylor expansion for the term $\Vert \tfrac{1}{N}\tilde{\vw} \Vert_2^{-1} = \Vert \tfrac{1}{N} (\vw+\tilde{\Delta})\Vert_2^{-1}$ up to the first order, which yields
\begin{equation}
\label{appendix-eq:16}
    \vw^{\text{next}} = \vw - \tfrac{1}{N}\tilde{\Delta}\tilde{\Delta}^\T \vw
    -\tfrac{1}{N}\vw(\vw^\T\tilde{\Delta}+\tfrac{1}{2}\tilde{\Delta}^\T\tilde{\Delta}) + \tilde{\Delta} + \delta,
\end{equation}
where $\delta$ includes all higher order terms. Note that $\tfrac{1}{N}\vw^\T\tilde{\Delta}= \tfrac{1}{N}\sum_{i=1}^N w_{i}\mathbb{E}[\tilde{\Delta}_{i}]+o(\tfrac{1}{N})$, $\tfrac{1}{N}\tilde{\Delta}^\T\tilde{\Delta}= \tfrac{1}{N} \sum_{i=1}^N \mathbb{E}[\tilde{\Delta}_{i}^2] + o(\tfrac{1}{N})$.
Substituting the mean and variance results of $\tilde{\Delta}_{i}$ into (\ref{appendix-eq:16}), we get
\begin{equation}
\label{appendix-eq:19}
\begin{aligned}
    \mathbb{E}[w^{\text{next}}_i-w_i] &= \tfrac{1}{N}w_i[q^\T g_i+\tau \vw^\T\phi(\vw) - \tfrac{N}{2} \langle \tilde{\Delta}_{i}^2\rangle] - \tfrac{1}{N}\mU_{i,:} \vg_i - \tfrac{\tau}{N}\phi(w_i)
    \\
    \mathbb{E}[(w^{\text{next}}_i-w_i)(w^{\text{next}}_j-w_j)] &= \langle \tilde{\Delta}_{i}\Tilde{\Delta}_{j} \rangle + o(\tfrac{1}{N}),
\end{aligned}
\end{equation}
where as defined in (\ref{appendix-eq:14})
\begin{equation*}
    \begin{aligned}
    \vg_i &= \tfrac{\tau}{2} \vq (\langle f^\prime_{12,i}\rangle+\langle f^\prime_{21,i}\rangle)  -  \tfrac{\tau}{2}\langle\vc f_{12,i}+\vc f_{21,i}\rangle  .
    \end{aligned}
\end{equation*}

\subsection{Derive PDE from Moments Estimation}
We characterize the asymptotic dynamics of the online learning algorithm when $N$ goes to infinity. First, we define the joint empirical measure of each row of the feature vector $\vu \in \mathbb{R}^{1 \times d_1} $ and its estimate $w\in \mathbb{R}$ as
\begin{equation}
    \mu_{t}^N(w,\vu) \overset{\text{def}}{=} \tfrac{1}{N} \sum_{i=1}^{N} \delta(w - w_i,\vu - \vu_{i}),
\end{equation}
where the continuous time  $t$ is a piece-wise interpolation of iteration step $k$ by $k = \lfloor tN \rfloor$, and $\vu_i$ is the $i$th row of the feature matrix $\mU$, and $w_i$ is the $i$th entry of the model weight vector $\vw$.

When $N \to \infty$,  each row of $\vw$  evolves independently. As a result, when $N \to \infty$, the sequence of random probability measures $\{\mu^N_t\}$ converges weakly to a deterministic measure $\mu$. Let $P_t(\vu,w)$ be the density function of the limiting measure $\mu_t(\vu,w)$ at time $t$. The stochastic process can be describe by a McKean-Vlasov  Equation
\begin{equation}
\label{appendix-eq:20}
    \partial_t P_t(\vu,w) = - \partial_{w} [\Gamma P_t(\vu,w)]
    +\tfrac{1}{2}  \partial^2_{w}[\Lambda P_t(\vu,w)],
\end{equation}
where $ \partial^2_{w}$ denotes the second order partial derivative with respect to $w$.
The drift and diffusion coefficient functions $\Gamma$ and $\Lambda$ can be derived from mean and variance of increment   $\Delta_{i}$ of the model weights in one-step update by
\begin{equation}
\begin{aligned}
    \mathbb{E}[w^{\text{next}}-w] &= \tfrac{1}{N}\Gamma + o(\tfrac{1}{N})
    \\
    \mathbb{E}[(w^{\text{next}}-w)^2] &=  \tfrac{1}{N}\Lambda+o(\tfrac{1}{N}).
\end{aligned}
\end{equation}
 Note that for difference entry index $i$ in the weight vector $\vw$, it has the same mean and variance given $w_i$ and $\vu_i$, thus we can remove the index $i$, and the drift and diffusion terms for the $w_i$ and $\vu_i$ valued as $w$ and $\vu$  become
\begin{equation}
\label{appendix-eq:23}
\begin{aligned}
    \Gamma(w,\vu,q_t, r_t) &= w[\vq^\T \vg+\tau r - \tfrac{1}{2}\Lambda] - \vu \vg-\tau\phi(w)
\\
     \Lambda(w,\vu,q_t, r_t)  &= 
         \tfrac{\tau^2}{4m} \Big[  \big(1+\langle (\gamma^{(2)})^2\rangle \big)\langle f_{12}^2\rangle +
         \big(1+\langle (\gamma^{(1)})^2\rangle\big) \langle f_{21}^2 \rangle + 2\big(1+\langle \gamma^{(1)}\gamma^{(2)}\rangle \big) \langle f_{12} f_{21}\rangle    \Big],
    \end{aligned}
\end{equation}
respectively, and 
\begin{equation}
\label{appendix-eq:24}
    \begin{aligned}
        \vq_t &= \textstyle \iint \vu^\T w P_t\text{d}w\text{d}\vu,
        \quad
        r_t = \textstyle \iint w \phi(w) P_t\text{d}w\text{d}\vu,
    \end{aligned}
\end{equation}
with the symbols $\vg$, $f$, and $f^{\prime}$ are  shorthand for 
\begin{equation}
\label{appendix-eq:25}
    \begin{aligned}
    \vg &= \tfrac{\tau}{2} \vq (\langle f^\prime_{12}\rangle+\langle f^\prime_{21}\rangle)  -  \tfrac{\tau}{2}\langle\vc f_{12}+\vc f_{21}\rangle  
     \\
     f_{\ell,\tilde{\ell}} &= -2 \big(\sigma(\Theta_\ell)-\bar{y}^{(\ell)} \big) \sigma^\prime (\Theta_{\tilde{\ell}} ),
     \quad \bar{y}^{(\ell)} = \langle \sigma( \Theta_{\ell} ) \rangle
     \\
     f^\prime_{\ell,\tilde{\ell}} &= \big(\tfrac{\partial}{\partial \Theta_{1}} + \tfrac{\partial}{\partial \Theta_{2}} \big) f_{\ell,\tilde{\ell}} 
     \\
     \Theta_{\ell} &= \vc^\T\vq + \va \sqrt{1-\vq^\T\vq}  + \vct{\gamma}^{(\ell)}, \; \ell=1,2.
    \end{aligned}
\end{equation}

We finish  the derivations of PDE in  Theorem 1.

\section{Detailed derivations for Theorem 2}
\label{appendix-sec:2}
In this section we will derive the ODE for $\vq_t$, when $\phi(x)$ is absent. 
When an L2 prior is imposed, $\phi(x)=x$, but due to normalization in  \eqref{appendix-eq:itr}, one will get the same limiting as  $\phi(x)=0$.

\begin{proof}[Proof of Theorem 2]
    According to (\ref{appendix-eq:24}) in Theorem 1
\begin{equation*}
    \begin{aligned}
        \vq_t &= \textstyle \iint \vu^\T w P_t\text{d}w\text{d}\vu,
        \\
        r_t &= \textstyle \iint w\phi(w)  P_t\text{d}w\text{d}\vu.
        \\
    \end{aligned}
\end{equation*}
When $\phi(x) = 0$, $r_t = 0$ and the derivative of $\vq_t$ is
\begin{equation*}
        \tfrac{\text{d}\vq_t}{\text{d}t} = \tfrac{\text{d}}{\text{d}t}\textstyle \iint \vu^\T w P_t\text{d}w\text{d}\vu
        = \textstyle \iint \vu^\T w \partial_t P_t\text{d}w\text{d}\vu.
\end{equation*}
Substituting (\ref{appendix-eq:20}) into the above expression, we have
\begin{equation}
    \tfrac{\text{d} \vq_t }{\text{d} t}= - \iint \vu^\T w \partial_{w} [\Gamma P_t]\text{d}w\text{d}\vu
    +\tfrac{1}{2}\iint \vu^\T w  \Lambda \partial^2_{w}P_t\text{d}w\text{d}\vu ;
\end{equation}

The r.h.s. terms can be further simplified by integrated by part with the boundary conditions being 0 at infinity
\begin{equation}
\begin{aligned}    
    - \iint \vu^\T w \partial_{w} [\Gamma P_t]\text{d}w\text{d}\vu
    & = - \vu^\T w \Gamma P_t \vert^{+\infty}_{-\infty}
    + \iint \vu^\T \Gamma P_t\text{d}w\text{d}\vu
    \\
    &= \vq (\vq^\T\vg -\tfrac{1}{2}\Lambda) - \vg ;
    \\ 
    \tfrac{1}{2}\iint \vu^\T w  \Lambda \partial^2_{w}P_t\text{d}w\text{d}\vu
    &= 
    \tfrac{1}{2} \vu^\T w  \Lambda \partial_{w}P_t \vert_{-\infty}^{+\infty}
    -
    \tfrac{1}{2}\Lambda \iint \vu^\T   \partial_{w}P_t\text{d} w \text{d}\vu
    \\
    &= -
    \tfrac{1}{2}\Lambda \vu^\T   \partial_{w}P_t\vert^{+\infty}_{-\infty} 
    \\
    &= 0.
\end{aligned}
\end{equation}

Thus we get the ODE for $\vq_t$ 
\begin{equation}
\label{appendix-eq:28}
    \tfrac{\text{d}\vq}{\text{d}t} = \vq (\vq^\T\vg -\tfrac{1}{2}\Lambda) - \vg,
\end{equation}
where  $\vg$ and $\Lambda$ defined in  \eqref{appendix-eq:14} and \eqref{appendix-eq:23}, which are function of $\vq$. Thus, the above ODE is closed w.r.t. $\vq$.
\end{proof}

\section{Detailed derivation for Section \uppercase\expandafter{\romannumeral4}}
\label{appendix-sec:3}
In this section we present the derivations and experiments of Section \uppercase\expandafter{\romannumeral4} in the main text. 
In particular, we study how the hidden variable of feature and  additive noises in  data argumentation affect the training process. 

To simplify the analysis, in this section we remove the batch normalization and set $\bar{y}=0$. Numerical experiments shows that discarding this mean-removal operation reduces the number of terms in many equations of theoretical analysis, and it does not qualitatively change the property of the dynamics only slightly degrading the performance.

\subsection{Feature learnability depends on the second moment only}
\label{appendix-subsec:31}
\begin{proof}[Proof of Proposition 3]
In $d_1=1$ case, the ODE in (\ref{appendix-eq:28}) is

\begin{equation*}
    \tfrac{\text{d}q}{\text{d}t} = q (q g -\tfrac{1}{2}\Lambda) - g, 
\end{equation*}
where the diffusion and drift terms from (\ref{appendix-eq:23}) and (\ref{appendix-eq:25}) in Theorem 1 reduce to
\begin{equation}
\label{appendix-eq:29}
    \begin{aligned}
        g(q) &= -\tau \langle c \sigma(\Theta_1) \sigma'(\Theta_2)+c\sigma(\Theta_2) \sigma'(\Theta_1)\rangle + \tau q \langle \sigma(\Theta_1)\sigma''(\Theta_2)+\sigma(\Theta_2)\sigma''(\Theta_1)+2\sigma'(\Theta_1)\sigma'(\Theta_2) \rangle
        \\
        \Lambda &=\tfrac{\tau^2}{m} \big[  (\langle (\gamma^{(2)})^2 \rangle + 1) \langle \sigma(\Theta_1)^2\sigma(\Theta_2)'^2\rangle +(\langle (\gamma^{(1)})^2 \rangle + 1)\langle \sigma(\Theta_2)^2\sigma(\Theta_1)'^2\rangle   
        \\
        &+2(1+\langle \gamma^{(1)} \gamma^{(2)} \rangle) \langle \sigma(\Theta_1) \sigma(\Theta_1)' \sigma(\Theta_2) \sigma(\Theta_2)'\rangle \big],
    \end{aligned}
\end{equation}
in which $\Theta_{1} =cq +  \sqrt{1-q^2} e  +\gamma^{(1)}$ and $\Theta_{2} =cq + \sqrt{1-q^2} e +\gamma^{(2)}$, and we directly set $\bar{y}=0$ to simplify the expression.  For the convenience of analysis, we define a function of $q$
\begin{equation}
\label{appendix-eq:30}
    h(q) = q (q g(q) -\tfrac{1}{2}\Lambda(q)) - g(q).
\end{equation}

When $q = 0$, we get
\begin{equation*}
    \begin{aligned}
    h(0) &= -g(0)
    \\
    &= \tau \langle c \sigma(e  +\gamma^{(1)}) \sigma'(e  +\gamma^{(2)})+c \sigma(e  +\gamma^{(2)}) \sigma'(e  +\gamma^{(1)})\rangle
    \\
    &= 0.
    \end{aligned}
\end{equation*}
Thus, the initial state at $q=0$ is a fixed point of  the ODE (\ref{appendix-eq:30}). 

To analysis the stability of the fixed point at $q=0$, we investigate the first order of  the derivative of $h(q)$ at $q=0$. The ODE is stable if the first-order derivative is negative.  
For abbreviations, denote $\sigma_1 = \sigma(e +\gamma^{(1)}),\sigma_2 = \sigma(e +\gamma^{(2)})$. Using the Taylor expansion around $q=0$ up to the first order of $q$, each part of (\ref{appendix-eq:29}) becomes
\begin{equation*}
    \begin{aligned}
        \sigma(\Theta_1)& = \sigma_1 + cq\sigma'_1 + o(q)
        \\
        \sigma'(\Theta_1) &= \sigma'_1 + cq\sigma''_1+ o(q)
        \\
        \sigma''(\Theta_1) &= \sigma''_1 + cq\sigma'''_1+ o(q)
        \\
        \langle c\sigma(\Theta_1) \sigma'(\Theta_2)\rangle& = \langle c 
        (\sigma_1 \sigma_2' + cq(\sigma_1 \sigma_2'' + \sigma_1'\sigma_2'))
        \rangle+ o(q)
        = \langle c^2 \rangle q\langle \sigma_1 \sigma_2'' + \sigma_1'\sigma_2' \rangle+ o(q)
        \\
        \langle \sigma(\Theta_1) \sigma''(\Theta_2)\rangle&=\langle \sigma_1\sigma_2'' \rangle+ o(q)
        \\
        \langle \sigma'(\Theta_1) \sigma'(\Theta_2)\rangle&=\langle \sigma_1'\sigma_2' \rangle+ o(q)
        \\
    \langle \sigma^2(\Theta_1)\sigma'^2(\Theta_2) \rangle&=
    \langle  \sigma_1^2\sigma_2'^2 \rangle+ o(q)
     \\
    \langle \sigma(\Theta_1)\sigma'(\Theta_2)\sigma(\Theta_2)\sigma'(\Theta_1) \rangle&
    = \langle  \sigma_1 \sigma_1' \sigma_2 \sigma_2' \rangle+ o(q).
    \\
\end{aligned}
\end{equation*}
As $0< q\ll 1$, the functions in (\ref{appendix-eq:29}) are simplified as
\begin{equation}
    \begin{aligned}
        g(q) &= -\tau q (\langle c^2 \rangle-1)\langle \sigma_1 \sigma_2''+\sigma_2 \sigma_1'' + 2\sigma_1'\sigma_2' \rangle + o(q)
        \\
    \Lambda &= \tfrac{\tau^2}{m} \big[ (\langle (\gamma^{(2)})^2 \rangle + 1) \langle \sigma_1^2\sigma_2'^2 \rangle+(\langle (\gamma^{(1)})^2 \rangle + 1) \langle \sigma_2^2\sigma_1'^2 \rangle+  2(1+\langle \gamma^{(1)} \gamma^{(2)} \rangle)\langle \sigma_1 \sigma_1' \sigma_2 \sigma_2'\rangle \big] + o(q) ,
    \\
    \end{aligned}
\end{equation}
where $o(q)$ represents the higher order terms of $q$. Thus we get
\begin{equation*}
\begin{aligned}
    h(q) &= q(qg-\tfrac{1}{2}\Lambda) -g
    \\
    &= \tau  q (\langle c^2 \rangle-1)\langle \sigma_1 \sigma_2''+\sigma_2 \sigma_1'' + 2\sigma_1'\sigma_2' \rangle 
    \\
    &- q\tfrac{\tau^2}{2m} \big[ (\langle (\gamma^{(2)})^2 \rangle + 1) \langle \sigma_1^2\sigma_2'^2 \rangle+(\langle (\gamma^{(1)})^2 \rangle + 1) \langle \sigma_2^2\sigma_1'^2 \rangle+  2(1+\langle \gamma^{(1)} \gamma^{(2)} \rangle)\langle \sigma_1 \sigma_1' \sigma_2 \sigma_2'\rangle \big] + o(q).
    \\
\end{aligned}
\end{equation*}

Therefore, the derivative of $h(q)$ at $q=0$ is

\begin{equation}
    h'(q)\bigl|_{q=0} = \tau (\langle c^2 \rangle-1)\langle \sigma_1 \sigma_2'' + 2\sigma_1'\sigma_2'+ \sigma_2 \sigma_1''  \rangle - \tfrac{\tau^2}{2m} \big[  (\langle (\gamma^{(2)})^2 \rangle + 1) \langle \sigma_1^2\sigma_2'^2 \rangle +(\langle (\gamma^{(1)})^2 \rangle + 1) \langle \sigma_2^2\sigma_1'^2 \rangle +  2(1+\langle \gamma^{(1)} \gamma^{(2)} \rangle)\langle \sigma_1 \sigma_1' \sigma_2 \sigma_2'\rangle \big]
\end{equation}

The fixed point $q=0$ is unstable when $h'(q)\bigl|_{q=0}>0$ i.e. the model is trainable from $q=0$, thus we have the trainability condition for $\langle c^2\rangle$
\begin{equation}
    \langle c^2 \rangle > 1 + \tfrac{\tau}{2m} \big[  (\langle (\gamma^{(2)})^2 \rangle + 1) \langle \sigma_1^2\sigma_2'^2 \rangle +(\langle (\gamma^{(1)})^2 \rangle + 1) \langle \sigma_2^2\sigma_1'^2 \rangle +  2(1+\langle \gamma^{(1)} \gamma^{(2)} \rangle)\langle \sigma_1 \sigma_1' \sigma_2 \sigma_2'\rangle \big] /\langle \sigma_1 \sigma_2^{\prime \prime} + 2\sigma_1^{\prime} \sigma_2^{\prime}+ \sigma_2 \sigma_1^{\prime \prime}  \rangle.
\end{equation}
\end{proof}

This accounts for Proposition 3 in Section \uppercase\expandafter{\romannumeral4}. And for $d_1\geq 2$ case, we can analysis similarly by studying the Jacobi matrix around zero point $\vq = 0$.

\paragraph*{Example of the case with single data feature}

Here we use $\sigma(x)=x^2$ as the activation function so that the ODE expression can be simplified to a polynomial. Also we consider no additive noise. Thus the two branches are the same. 
\begin{claim}
In the setting with $d_1=1$ and with no additive noises, denoting $Q=q^2$, 
 the ODE for $Q$ is
\begin{equation}
\label{appendix-eq:34}
\begin{aligned}
    \tfrac{\text{d}Q}{\text{d}t} &= 
    8\tau (1-Q)(Q^2(m_4-3)+3Q(1-2Q)(m_2-1)) 
    \\&- 16 \tfrac{\tau^2}{m}Q(Q^3(m_6-15)+15Q^2(1-Q)(m_4-3)+45Q(1-Q)^2 (m_2-1)+15) ,
\end{aligned}
\end{equation}
where $m_2,m_4,m_6$ are the 2nd, 4th, 6th moments of $c$ respectively.
\end{claim}

\begin{figure}[htbp]

\subfloat[Gradient of $Q$]
 {
    \label{appendix-fig:subfig21}\includegraphics[width=0.44\columnwidth]{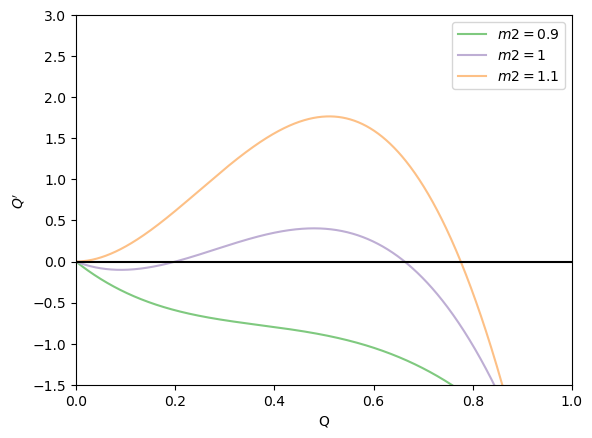}
 }\hfill
\subfloat[Training dynamics of $Q$]
 {
    \label{appendix-fig:subfig22}\includegraphics[width=0.53\columnwidth]{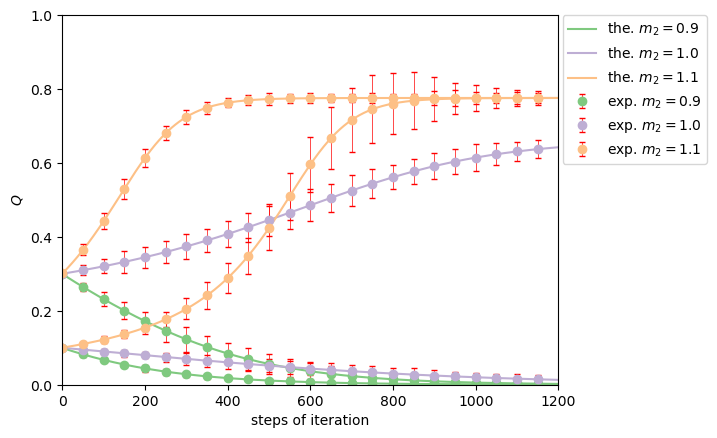}
 }

\caption{\label{appendix-fig:2} (a) Gradient of $Q$ with different $m_2$, (b) Training dynamics of $Q$ with different $m_2$.
}
\end{figure}

The experiment results are showed in Fig. \ref{appendix-fig:2}. The gradient of $Q$ with different $m_2$ is plotted in Fig. \ref{appendix-fig:subfig21}. 
When $m_2=1$. there are two fixed points. One is stable at $Q=0$, the other is stable near $q=1$. When $m_2 = 1.1 > m_{u}$ for some constant $m_{u}$, the fixed point $q=0$ becomes unstable. On the contrary, when $m_2 = 0.9 > m_{l}$ for some constant $m_{l}$, the fixed point near $q=1$ vanishes and the only fixed point $q=0$ remain stable. This insight agrees that the trivial, non-informative solution have less chance to be reached as $m_2$ increases.
 The corresponding training dynamics are showed in Fig. \ref{appendix-fig:subfig22}. For the starting point $Q=0.1$, only feature with $m_2=1.1$ can be retrieved.  If initialized from $q=0.3$, only feature with $m_2=0.9$ is not retrieved. This results shows that $m_2$ also affect the attraction region of the training dynamics.  

\begin{proof}[Proof of Claim 1]
Because no additive noise is considered, we can omit the subscript for different branches in (\ref{appendix-eq:25}) and get
    \begin{equation*}
    \begin{aligned}
        f(\Theta) &= -4\Theta^3 
        \\
        f'(\Theta) &= -12\Theta^3,
    \end{aligned}
\end{equation*}
where $\Theta = cq + \sqrt{1-q^2}e$, we denote the moments of $c$ as $m_2,m_4,m_6$, we can calculate
\begin{equation*}
    \begin{aligned}
        \langle \Theta^6 \rangle &= \langle  (cq + \sqrt{1-q^2}e)^6 \rangle
        \\
        &=q^6m_6 + 15q^4 (\sqrt{1-q^2}) ^2m_4 + 45q^2 (\sqrt{1-q^2})^4 m_2 + 15 (\sqrt{1-q^2})^6
        \\
        &= q^6m_6 + 15q^4 (1-q^2)m_4 + 45q^2 (1-2q^2+q^4)m_2 +15 (1-3q^2+3q^4-q^6)
        \\
        &= q^6(m_6-15)+15q^4(1-q^2)(m_4-3)+45q^2(1-q^2)^2 (m_2-1) +15
        \\
           \langle c\Theta^3 \rangle&= \langle c(cq + \sqrt{1-q^2}e)^3 \rangle 
           \\
            &=\langle c (cq)^3 + 3c^2q\sqrt{1-q^2}^2\rangle
           \\
           &=  q^3(m_4 -3) + 3q(1-q^2)   (m_2-1) + 3q
           \\
            \langle \Theta^2 \rangle&= \langle (cq + \sqrt{1-q^2}e)^2 \rangle 
           \\
           &=  m_2 q^2  
           + (1-q^2) =1 + q^2(m_2-1)     
           \\
           \langle c\Theta^3 - 3q\Theta^2\rangle&= q^3(m_4-3)+3q(1-2q^2)(m_2-1).
    \end{aligned}
\end{equation*}

Thus the expressions of $g$ and $\Lambda$ are

\begin{equation}
    \begin{aligned}
        g&= -4\tau[q^3(m_4-3)+3q(1-2q^2)(m_2-1)]
        \\
        \Lambda &= \tfrac{16\tau^2}{m}[q^6(m_6-15)+15q^4(1-q^2)(m_4-3)+45q^2(1-q^2)^2 (m_2-1) +15].
    \end{aligned}
\end{equation}

According to Theorem 2, the ODE is
\begin{equation}
\label{appendix-eq:36}
\begin{aligned}
    \tfrac{\text{d}q}{\text{d}t} 
    &= q (qg - \tfrac{1}{2} \Lambda) - g
    \\
    &= (q^2-1)g - \tfrac{1}{2}q\Lambda
    \\
    &=
4\tau (1-q^2)(q^3(m_4-3)+3q(1-q^2)(1-2q^2)(m_2-1)) 
\\
&- 8\tfrac{\tau^2}{m}q(q^6(m_6-15)+15q^4(1-q^2)(m_4-3)+45q^2(1-q^2)^2 (m_2-1)+15) .
\end{aligned}
    \end{equation}

We define $Q=q^2$ and notice that
\begin{equation*}
    2q\tfrac{\text{d}q}{\text{d}t} = \tfrac{\text{d}Q}{\text{d}t},
\end{equation*}
combined with (\ref{appendix-eq:36}), we get (\ref{appendix-eq:34}) in Claim 1.
\end{proof}

\subsection{Higher moments affect feature selection probability as well}
\label{appendix-subsec:32}
In this subsection, we  investigate the effects of data model with two features, {\em i.e.} $d_1=2$. We restrict the setting that  no additive noises is presented, and  the activation function is $\sigma(x)=x^2$, in order to get a polynomial expression of ODE.

The two branch of the contrastive model become the same when there is no additive noise. Thus, we omit the subscript for different branches in (\ref{appendix-eq:25}) and get
\begin{equation}
    \begin{aligned}
        f(\Theta) &= -4\Theta^3
        \\
        f'(\Theta) &= -12\Theta^2,
    \end{aligned}
\end{equation}
where $
\Theta =  \vc^\T\vq + \sqrt{1-\vq^\T\vq} e        
$, here $\vc\in \mathbb{R}^{2},\vq\in\mathbb{R}^2$.

\begin{claim}
Define the 2nd, 4th, 6th moments of the hidden variables $\vc$ respectively as $m_2^1,m_4^1,m_6^1$ for $c_1$ and $m_2^2,m_4^2,m_6^2$ for $c_2$. For $Q_1=q_1^2,Q_2=q_2^2$, they satisfy the follow ODEs
\begin{equation}
\label{appendix-eq:38}
\begin{aligned}
    \tfrac{\text{d}Q_1}{\text{d}t} &= 
     4\tau \big[\kappa_1 Q_1^2 + 3 \nu_1Q_1(1-2Q_1)  + 3\nu_1\nu_2Q_1Q_2 \big](1-Q_1)
            -4\tau \big[\kappa_2 Q_2 + 3 \nu_2(1-2Q_2^2) +3\nu_1\nu_2Q_1 \big]Q_1Q_2
            \\
            &-\tfrac{8\tau^2}{m}Q_1 \Big[\beta_1Q_1^2 +\beta_2Q_2^3 +15\kappa_1Q_1^2(1-Q_1)+15\kappa_2Q_2^2(1-Q_2)
        +45\nu_1Q_1(Q_1^2-2Q_1+1) +45\nu_2Q_2(Q_2^2- 2 Q_2 + 1) 
        \\
        &+15\kappa_1\nu_2Q_1^2Q_2 + 15\kappa_2\nu_1Q_1Q_2^2
        + 90\nu_1\nu_2Q_1Q_2(1-Q_1-Q_2)+ 15 \Big]
        \\
        \tfrac{\text{d}Q_2}{\text{d}t} &= 
        \text{switch the subscripts 1,2 of all symbols in r.h.s of the above equation.}
\end{aligned}
\end{equation}
where $\nu_1=m_{2}^1-1=,\kappa_1=m_{4}^1-3,\beta_1=m_{6}^1-15$ and $\nu_2=m_{2}^2-1,\kappa_2=m_{4}^2-3,\beta_2=m_{6}^2-15$.
\end{claim}

\begin{figure}[htbp] 

\centering
\subfloat[$Q_1'$ in case one]
  {
    \label{appendix-fig:subfig31}\includegraphics[width=0.32\columnwidth]{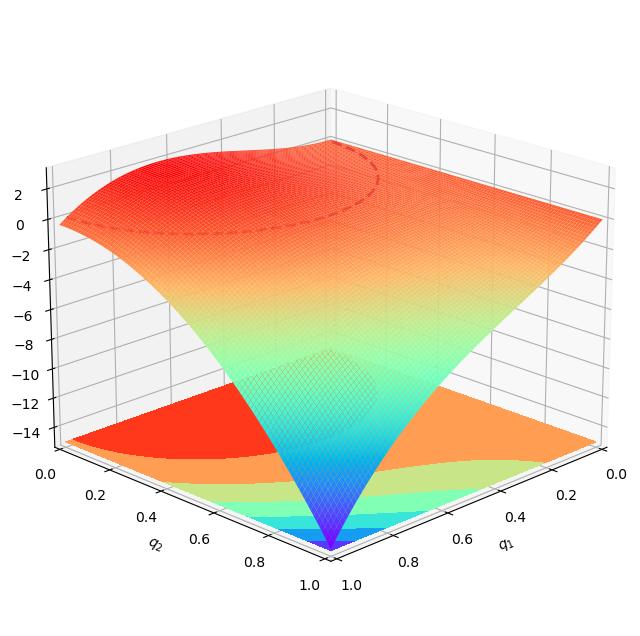}
  }
  \subfloat[$Q_2'$ in case one]
  {
    \label{appendix-fig:subfig32}\includegraphics[width=0.32\columnwidth]{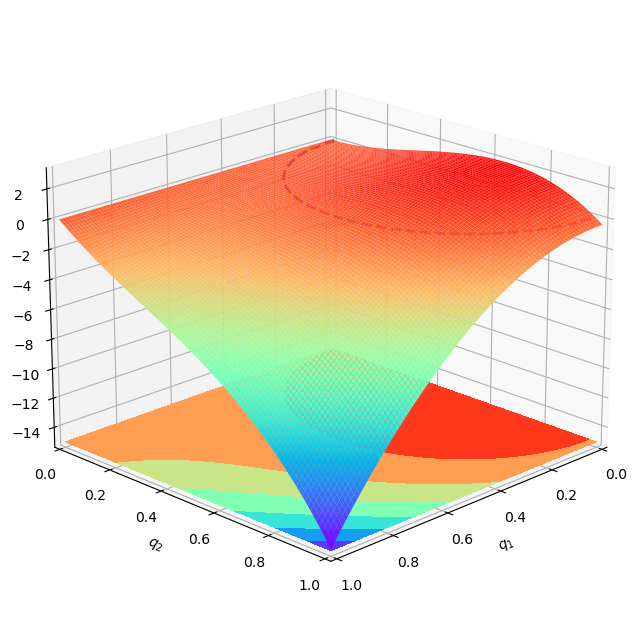}
  }
  \subfloat[$Q_2'-Q_1'$ in case one]
  {
    \label{appendix-fig:subfig33}\includegraphics[width=0.32\columnwidth]{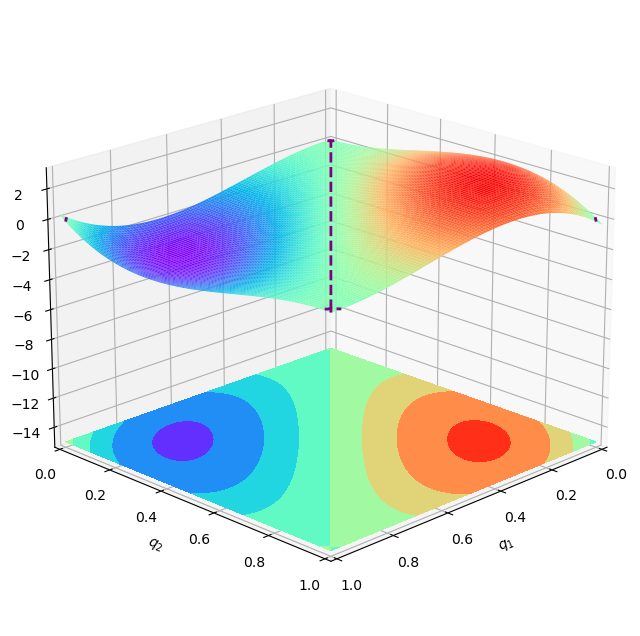}
  }\hfill
  \\
  \subfloat[$Q_1'$ in case two]
  {
    \label{appendix-fig:subfig34}\includegraphics[width=0.32\columnwidth]{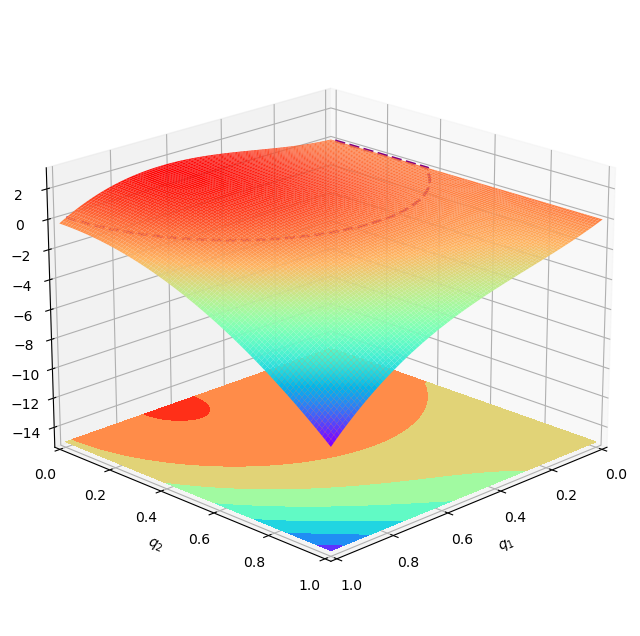}
  }
  \subfloat[$Q_2'$ in case two]
  {
    \label{appendix-fig:subfig35}\includegraphics[width=0.32\columnwidth]{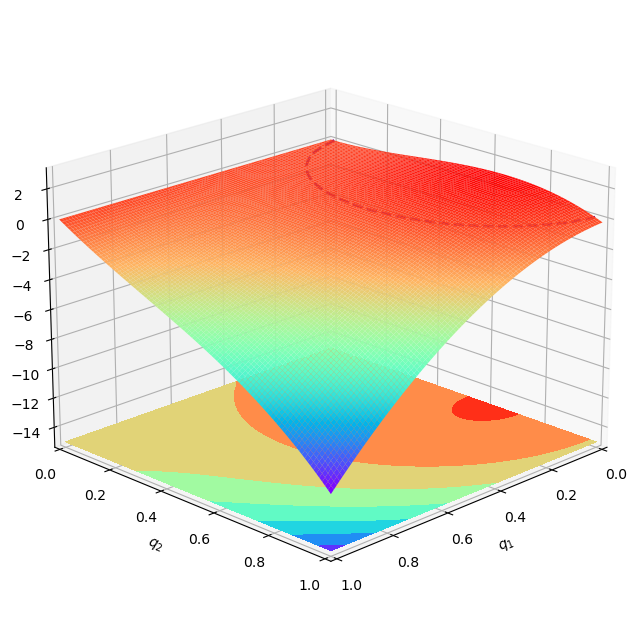}
  }
  \subfloat[$Q_2'-Q_1'$ in case two]
  {
    \label{appendix-fig:subfig36}\includegraphics[width=0.32\columnwidth]{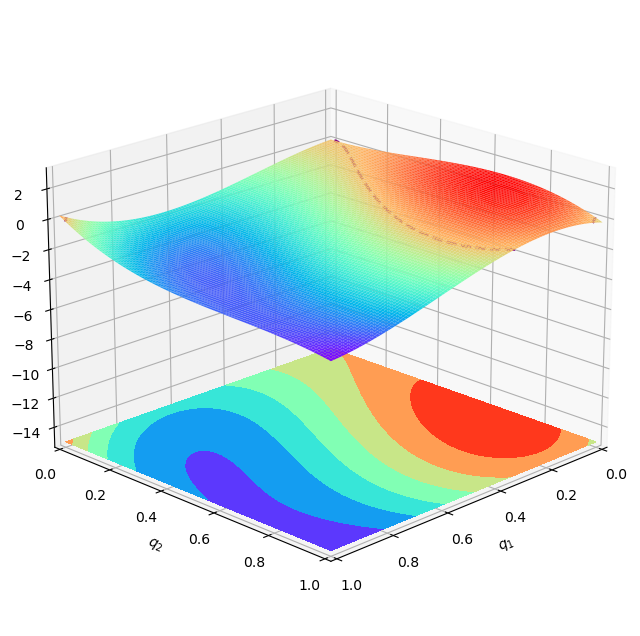}
  }\hfill
\caption{Gradient in two cases: (a-c) $c_1$ and $c_2$ both obey Gaussian distribution with $m_2^1 = m_2^2 = 1$ and $m_4^1 = m_4^2 = 3,m_6^1 = m_6^2 = 15$.  (d-f) $c_1$ obey Gaussian distribution with $m^1_2=1.2,m^1_4=4.32,m^1_6=25.95$ while $c_2$ obey Three-points distribution with $m_2^2=1.1$ and $m_4^2=6.05,m_6=33.275$.}
\label{appendix-fig:3}
\end{figure}

The gradients in symmetrical and asymmetrical cases are plotted in Fig. \ref{appendix-fig:3}. The hidden variables $c_1$ and $c_2$ obey same Gaussian distribution in symmetrical case. In the asymmetrical case, $c_1$ obeys Gaussian distribution and $c_2$ obeys Three-points distribution which means $c_2 \in \{\pm \sqrt{\alpha},1\}$ with probability $P(c_2=\sqrt{\alpha}) = P(c_2=\sqrt{\alpha}) = p/2,P(c_2=0) = 1-p,$ where $p,\alpha$ are parameters of the distribution. 

\begin{figure}[t] 

\centering
\subfloat[Phase portrait in sym. case]
  {
    \label{appendix-fig:subfig41}\includegraphics[width=0.25\columnwidth]{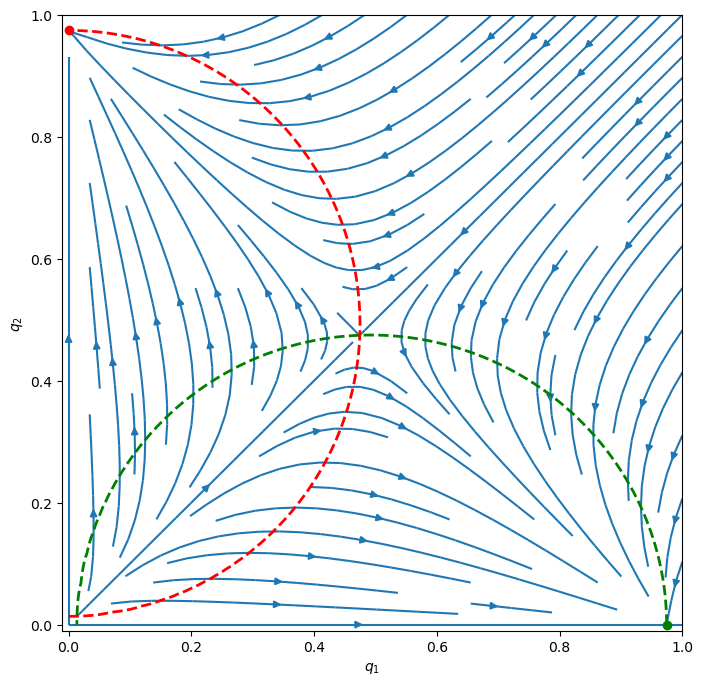}
  }
  \subfloat[Phase portrait in sym. case]
  {
    \label{appendix-fig:subfig42}\includegraphics[width=0.25\columnwidth]{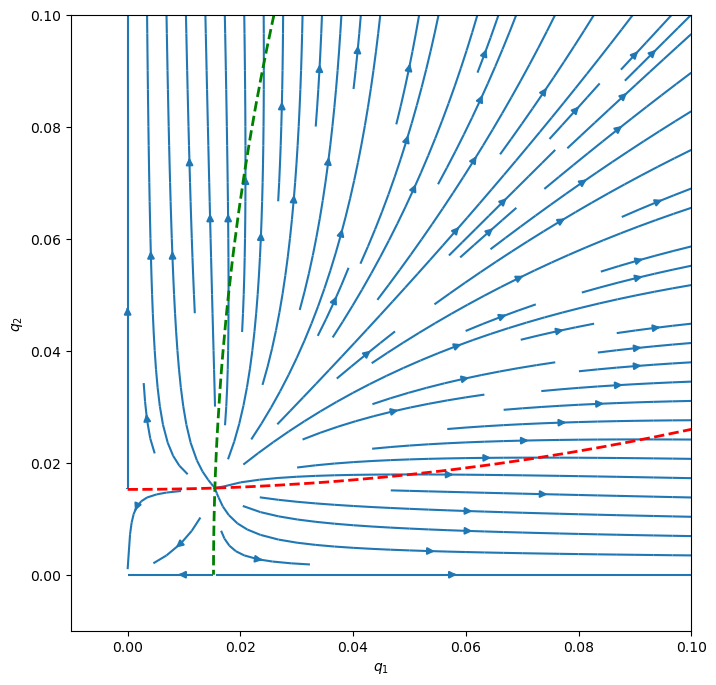}
  }
  \subfloat[Phase portrait in sym. case]
  {
    \label{appendix-fig:subfig43}\includegraphics[width=0.25\columnwidth]{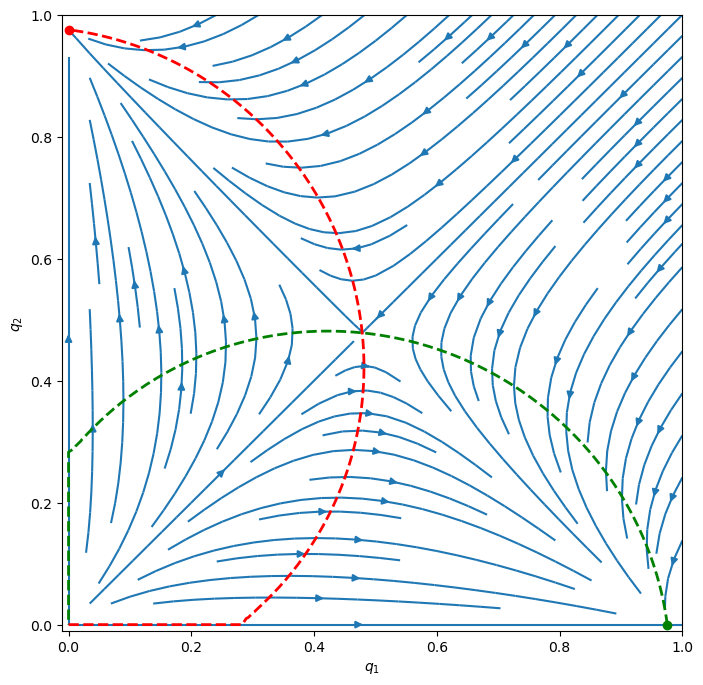}
  }\hfill
\caption{Phase portrait in symmetrical cases: (a-b) $c_1$ and $c_2$ both obey Gaussian distribution with $m_2^1 = m_2^2 = 1$ and $m_4^1 = m_4^2 = 3,m_6^1 = m_6^2 = 15$.  (c) $c_1$ and $c_2$ both obey Gaussian distribution with $m_2^1 = m_2^2 = 1.1$ and $m_4^1 = m_4^2 = 3.63,m_6^1 = m_6^2 = 19.965$.}
\label{appendix-fig:4}
\end{figure}

\begin{figure}[t] 

\centering
\subfloat[Phase portrait in asy. case]
  {
    \label{appendix-fig:subfig51}\includegraphics[width=0.25\columnwidth]{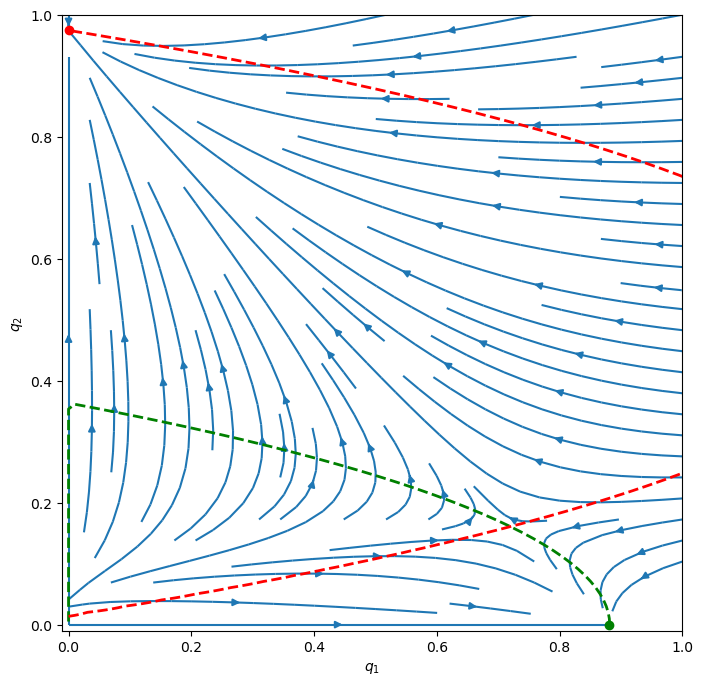}
  }
  \subfloat[Phase portrait in asy. case]
  {
    \label{appendix-fig:subfig52}\includegraphics[width=0.25\columnwidth]{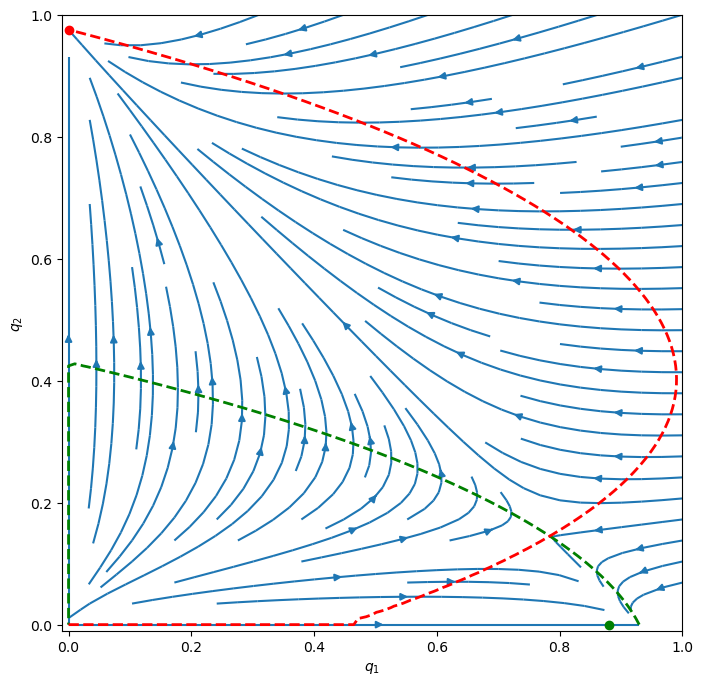}
  }\hfill
\caption{Phase portrait in asymmetrical cases: (a) $c_1$ obey Gaussian distribution with $m^1_2=1.1,m^1_4=3.63,m^1_6=19.965$ while $c_2$ obey Three-points distribution with $m_2^2=1$ and $m_4^2=5,m_6=25$.  (b) $c_1$ obey Gaussian distribution with $m^1_2=1.2,m^1_4=4.32,m^1_6=25.95$ while $c_2$ obey Three-points distribution with $m_2^2=1.1$ and $m_4^2=6.05,m_6=33.275$.}
\label{appendix-fig:5}
\end{figure}

 Fig. \ref{appendix-fig:4} and Fig.  \ref{appendix-fig:5} show the phase portrait visualizing the fixed points and convergence region. The dotted line represents the contour of zero points.  In symmetrical case, there are four fixed points in total. The fixed point near $Q_1,Q_2=(0,0)$ is stable in $m_2^1=m_2^2=1.0$ case while it is unstable in $m_2^1=m_2^2=1.1$ case. The fixed point near $Q_1,Q_2=(0.5,0.5)$ is unstable in both cases meaning only one feature can be retrieved regardless of the second order moment. The remaining two fixed points near $(1,0)$ and $(0,1)$ are stable in both cases indicating that one of the two features is well retrieved. 

Similarly, in asymmetrical cases, four fixed points remain. The fixed point near $(0,0)$ is unstable because $m_2^1>1$ and the fixed point near $Q_1,Q_2=(0.5,0.5)$ is unstable as well. For the two fixed points near $(1,0)$ and $(0,1)$, the convergence region of feature 2 is bigger than that of feature 1. Since starting at a unstable fixed point at random initial state around $Q_1=Q_2=0$, which training trajectory will be selected is a random event, and its probability is affect by the portion of attraction region to different feature recovery state. Therefore, higher moments affect feature selection probability as well.

\begin{proof}[Proof of Claim 2]
The diffusion and drift terms from (\ref{appendix-eq:23}) and (\ref{appendix-eq:25}) in Theorem 1 are

\begin{equation}
    \begin{aligned}
    \vg &= -\tau \langle \vc f(\Theta)\rangle +\tau \vq\langle f'(\Theta)\rangle  \rangle = -4\tau \langle \vc \Theta^3 \rangle
    + 12\tau \vq \langle \Theta^2 \rangle
    \\
    \Lambda& = 
         \tfrac{\tau^2}{m}\langle f(\Theta)^2\rangle  = \tfrac{\tau^2}{m} 16 \langle \Theta^6\rangle.
         \end{aligned}
\end{equation}

We can calculate the expectation of $\Theta^k$
\begin{equation*}
    \begin{aligned}
           \langle \vc_1\Theta^3 \rangle&= \langle \vc^1(\vc^\T\vq + \sqrt{1-\vq^\T\vq}e)^3 \rangle 
           \\
            &=\langle \vc_1 (\vc^\T\vq)^3 + 3\vc\vc^\T\vq (\sqrt{1-\vq^\T \vq})^2\rangle
           \\
           &=  
                m_{4}^1q_1^3 + 3m_{2}^1 m_{2}^2q_1q_2^2 + 3 m_{2}^1 q_1 (1-\vq^\T\vq)  
           \\
           &=(m_4^1-3)q_1^3 + 3 (m_2^1-1)q_1(1-q_1^2-q_2^2)  + 3(m_2^1-1)(m_2^2-1)q_1 q_2^2
           +  3q_1  + 3(m_2^2+m_2^1-2)q_1q_2^2
           \\
           \langle \Theta^2 \rangle&= \langle (\vc^\T\vq + \sqrt{1-\vq^\T\vq}e)^2 \rangle 
           \\
            &=\langle (\vc^\T\vq)^2 + (\sqrt{1-\vq^\T\vq})^2\rangle
           \\
           &=  (m_{2}^1-1)q_1^2 + (m_{2}^2-1)q_2^2 + 1
           \\
            -\vg_1/4\tau &=  \langle \vc_1 \Theta^3 \rangle - 3q_1\langle \Theta^2\rangle 
           \\
           &=(m_4^1-3)q_1^3 + 3 (m_2^1-1)q_1(1- 2q_1^2-2q_2^2)  + 3(m_2^1-1)(m_2^2-1)q_1\vq^{2}_2
            \\
            -\vq^\T\vg/4\tau &=(m_{4}^1-3) q_1^4 + 3 (m_{2}^1-1)q_1^2(1-2q_1^2)  + 6(m_{2}^1-1)(m_{2}^2-1)q_1^2q_2^2
            +(m_{4}^2-3) q_2^4 + 3 (m_{2}^2-1)q_2^2(1-2q_2^2).
   \end{aligned}
\end{equation*}

The expectation in the variance term is
\begin{equation*}
    \begin{aligned}
        \langle \Theta^6 \rangle &= \langle  (\vc^\T\vq + \sqrt{1-\vq^\T\vq}e)^6 \rangle
        \\
        &=\langle (\vc^\T\vq)^6 \rangle + 15 \langle (\vc^\T\vq)^4\rangle (\sqrt{1-\vq^\T\vq}) ^2 + 45\langle (\vc^\T\vq)^2 \rangle (\sqrt{1-\vq^\T\vq})^4 + 15 (\sqrt{1-\vq^\T\vq})^6
        \\
        &= (m_{6}^1-15)q_1^6 +(m_{6}^2-15)q_2^6 +15(m_{4}^1-3)q_1^4(1-q_1^2)+15(m_{4}^2-3)q_2^4(1-q_2^2)
        \\
        &+45(m_{2}^1-1)q_1^2(q_1^4-2q_1^2+1) +45(m_{2}^2-1)q_2^2(q_2^4- 2 q_2^2 + 1) 
        \\
        &+15(m_{4}^1-3)(m_{2}^2-1)q_1^4q_2^2 + 15(m_{4}^2-3)(m_{2}^1-1)q_1^2q_2^4
        + 90(m_{2}^1-1)(m_{2}^2-1)q_1^2q_2^2(1-q_1^2-q_2^2)+ 15.
        \\
    \end{aligned}
\end{equation*}

Thus we have the drift and diffusion terms
\begin{equation}
    \begin{aligned}
        g_1 &= -4\tau[(m_4^1-3)q_1^3 + 3 (m_2^1-1)q_1(1- 2q_1^2-2q_2^2)  + 3(m_2^1-1)(m_2^2-1)q_1q_2^{2}]
        \\
        g_2 &= -4\tau[(m_4^2-3)q_2^3 + 3 (m_2^2-1)q_2(1- 2q_1^2-2q_2^2)  + 3(m_2^2-1)(m_2^1-1)q_2q_1^{2}]
        \\
        \Lambda &= \tfrac{16\tau^2}{m}[(m_{6}^1-15)q_1^6 +(m_{6}^2-15)q_2^6 +15(m_{4}^1-3)q_1^4(1-q_1^2)+15(m_{4}^2-3)q_2^4(1-q_2^2)
        \\
        &+45(m_{2}^1-1)q_1^2(q_1^4-2q_1^2+1) +45(m_{2}^2-1)q_2^2(q_2^4- 2 q_2^2 + 1) 
        \\
        &+15(m_{4}^1-3)(m_{2}^2-1)q_1^4q_2^2 + 15(m_{4}^2-3)(m_{2}^1-1)q_1^2q_2^4
        + 90(m_{2}^1-1)(m_{2}^2-1)q_1^2q_2^2(1-q_1^2-q_2^2)+ 15].
    \end{aligned}
\end{equation}

According to Theorem 2, the ODEs for $\vq$ are

    \begin{equation}
\begin{aligned}
    \tfrac{\text{d}q_1}{\text{d}t} &= q_1(\vq^\T\vg - \tfrac{1}{2}\Lambda) - g_1
    \\
    &= q_1(\vq^\T\vg) - g_1 - q_1\tfrac{1}{2}\Sigma
    \\
    &= 4\tau [(m_{4}^1-3) q_1^3 + 3 (m_{2}^1-1)q_1(1-2q_1^2)  + 3(m_{2}^1-1)(m_{2}^2-1)q_1q_2^2](1-q_1^2)
            \\
            &-4\tau[(m_{4}^2-3) q_2^3 + 3 (m_{2}^2-1)q_2(1-2q_2^2) +3(m_{2}^1-1)(m_{2}^2-1)q_1^2q_2]q_1q_2
            \\
            &-\tfrac{8\tau^2}{m}q_1[(m_{6}^1-15)q_1^6 +(m_{6}^2-15)q_2^6 +15(m_{4}^1-3)q_1^4(1-q_1^2)+15(m_{4}^2-3)q_2^4(1-q_2^2)
        \\
        &+45(m_{2}^1-1)q_1^2(q_1^4-2q_1^2+1) +45(m_{2}^2-1)q_2^2(q_2^4- 2 q_2^2 + 1) 
        \\
        &+15(m_{4}^1-3)(m_{2}^2-1)q_1^4q_2^2 + 15(m_{4}^2-3)(m_{2}^1-1)q_1^2q_2^4
        \\
        &+ 90(m_{2}^1-1)(m_{2}^2-1)q_1^2q_2^2(1-q_1^2-q_2^2)+ 15]
        \\
        \tfrac{\text{d}q_2}{\text{d}t} &= q_2(\vq^\T\vg - \tfrac{1}{2}\Lambda) - \vg_2
    \\
    &= \text{switch the subscripts 1,2 of all symbols in r.h.s of the above equation.}
\end{aligned}
    \end{equation}
switch the subscripts 1,2 of all symbols in r.h.s of the above equation.
Denote $q_1^2 = Q_1, q_2^2 = Q_2$, we have the ODEs for $Q_1,Q_2$
\begin{equation}
\begin{aligned}
    \tfrac{\text{d}Q_1}{\text{d}t} &= 
     4\tau [(m_{4}^1-3) Q_1^2 + 3 (m_{2}^1-1)Q_1(1-2Q_1)  + 3(m_{2}^1-1)(m_{2}^2-1)Q_1Q_2](1-Q_1)
            \\
            &-4\tau[(m_{4}^2-3) Q_2 + 3 (m_{2}^2-1)(1-2Q_2) +3(m_{2}^1-1)(m_{2}^2-1)Q_1]Q_1Q_2
            \\
            &-\tfrac{8\tau^2}{m}Q_1[(m_{6}^1-15)Q_1^3 +(m_{6}^2-15)Q_2^3 +15(m_{4}^1-3)Q_1^2(1-Q_1)+15(m_{4}^2-3)Q_2^2(1-Q_2)
        \\
        &+45(m_{2}^1-1)Q_1(Q_1^2-2Q_1+1) +45(m_{2}^2-1)Q_2(Q_2^2- 2 Q_2 + 1) 
        \\
        &+15(m_{4}^1-3)(m_{2}^2-1)Q_1^2Q_2 + 15(m_{4}^2-3)(m_{2}^1-1)Q_1Q_2^2
        \\
        &+ 90(m_{2}^1-1)(m_{2}^2-1)Q_1Q_2(1-Q_1-Q_2)+ 15]
        \\
        \tfrac{\text{d}Q_2}{\text{d}t} &= \text{switch the subscripts 1,2 of all symbols in r.h.s of the above equation.}
\end{aligned}
\end{equation}

For simplicity define $\nu_1 = m_{2}^1-1,\kappa_1=m_{4}^1-3 ,\beta_1=m_{6}^1-15$ and $\nu_2=m_{2}^2-1,\kappa_2=m_{4}^2-3 ,\beta_2=m_{6}^2-15$, we can get (\ref{appendix-eq:38}) finally.
\end{proof}

\subsection{Correlated additive noise reduces gradient variance}
\label{appendix-subsec:33}
In this part, we will focus on the role of the additive noise and supplement the theoretical derivation in Section \uppercase\expandafter{\romannumeral4} of the main text. 

We consider two types of additive noise, and study how they affect the training process by deriving the corresponding limiting ODE and compare the results with no additive noise case. For the convenience of analysis, we use $\sigma(x) = x^2$ as activation function and $d_1=1,\phi=0$ case is considered. $m_2,m_4,m_6$ are the 2nd, 4th, 6th moments of $c$ respectively, and $m_2$ is fixed i.e. $m_2=1$to concentrate on the impact of additive noise.
\paragraph{Independent Gaussian Noise}
\begin{claim}
If the additive noise of two branches is independent Gaussian noise, $\gamma^{(1)},\gamma^{(2)}$ are i.i.d. random variables sampling from $\mathcal{N}(0,\eta)$. In $d_1=1,\sigma(x)=x^2$ case, the subtraction of Gradient for $\eta >0$ and $\eta =0$ is

\begin{equation}
\label{appendix-eq:46}
        \begin{aligned}
    \tfrac{\text{d}Q}{\text{d}t} \bigg|_{\eta>0}- \tfrac{\text{d}Q}{\text{d}t} \bigg|_{\eta=0} 
    &=
    -\tfrac{4\tau^2}{m}Q[\eta(Q^3(m_6-15) + 15Q^2(1-Q)(m_4-3))  
    \\
    &\quad + 
    ( 7\eta^2 + 14\eta)(Q^2(m_4-3) + 3)
    +3\eta^4 + 12\eta^3 + 18\eta^2 
    ],
\end{aligned}
\end{equation}
where $Q = q^2$, $m_4,m_6$ are the 4th, 6th moments of $c$.
\end{claim}

\begin{figure}[t] 

\centering
\subfloat[Gradient of independent additive noise case]
  {
    \label{appendix-fig:subfig61}\includegraphics[width=0.45\columnwidth]{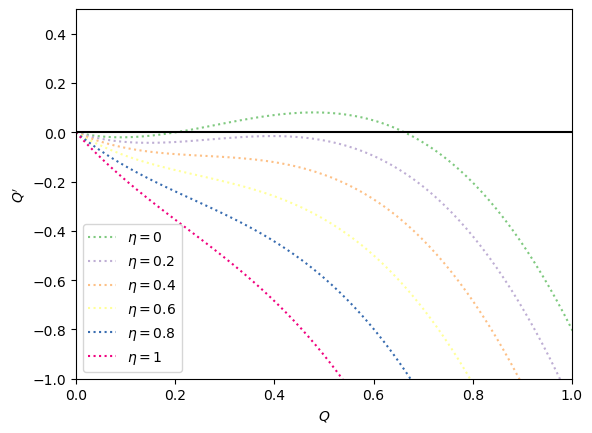}
  }
  \subfloat[Gradient of correlated additive noise case]
  {
    \label{appendix-fig:subfig62}\includegraphics[width=0.45\columnwidth]{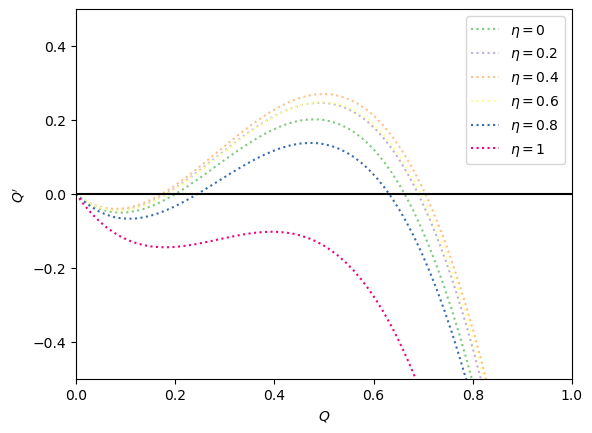}
  }\hfill
\caption{Gradient in independent and correlated additive Gaussian noise cases.}
\label{appendix-fig:6}
\end{figure}

When $m_4\geq 3,m_6\geq 15,\quad \tfrac{\text{d}Q}{\text{d}t} \big|_{_\eta>0}- \tfrac{\text{d}Q}{\text{d}t} \big|_{\eta=0}\leq0, \forall Q\in[0,1]$. It indicates that independent additive Gaussian noise has no positive impact. The experiment results are plotted in Fig. \ref{appendix-fig:subfig61}. As $\eta$ grows from $0$ to $1$, the gradient monotonically decreases at any $Q\in [0,1]$ meaning that independent additive Gaussian noise is harmful to feature retrieval. The fixed point near $Q=1$ vanishes when $\eta>0.2$. This observation can be verified theoretically from  (\ref{appendix-eq:46}).

\paragraph{Correlated Gaussian Noise}
\begin{claim}
If the additive noise of two branches is correlated Gaussian noise, $\gamma^{(1)}$ are random variables sampling from $\mathcal{N}(0,\eta)$ and $\gamma^{(2)} = -\gamma^{(1)}$. In $d_1=1,\sigma(x)=x^2$ case,  denote $Q=q^2$, the subtraction of gradient for $\eta>0$ and $\eta=0$ is
    \begin{equation}
    \label{appendix-eq:50}
    \tfrac{\text{d}Q}{\text{d}t} \bigg|_{\eta>0}- \tfrac{\text{d}Q}{\text{d}t} \bigg|_{\eta=0} 
    = 
    \tfrac{64\tau^2}{m}Q(Q^2(m_4-3) + 3)\eta+o(\eta).
\end{equation}
when $\eta$ is relatively small i.e. $\eta \ll 1$ and $o(\eta)$ includes all high order terms.

\end{claim}

The subtraction is positive when $m_4>3$ at any $Q\in[0,1]$. This confirms the results of the claim in section \uppercase\expandafter{\romannumeral4}.C of the main text that correlated additive noise reduces gradient variance and is beneficial to feature retrieval.
The gradient is plotted in Fig. \ref{appendix-fig:subfig62}. The gradient first increases and then decreases as $\eta$ grows from $0$ to $1$. As the gradient increases, the convergence region of $Q=0$ shrinks while the convergence region of $Q\to 1$ expands and the convergence value increases as well. The convergence values are showed explicitly in Fig. 4 in the main text.

\subsection*{Proofs of Claim 3 and Claim 4}

According to Theorem 1, the drift and diffusion terms are

\begin{equation}
\label{appendix-eq:45}
\begin{aligned}
    \Gamma &= w[\vq^\T \vg - \tfrac{1}{2}\Lambda] - \vu \vg
\\
     \Lambda &= 
         \tfrac{\tau^2}{4m} \Big[  \big(1+\langle (\gamma^{(2)})^2\rangle \big)\langle f_{12}^2\rangle + \big(1+\langle (\gamma^{(1)})^2\rangle\big) \langle f_{21}^2 \rangle + 2\big(1+\langle \gamma^{(1)}\gamma^{(2)}\rangle \big) \langle f_{12} f_{21}\rangle    \Big],
    \end{aligned}
\end{equation}

For activation function $\sigma(x) = x^2$, $g,f,f'$ can be calculated as
\begin{equation}
    \begin{aligned}
    g &= \tfrac{\tau}{2} q (\langle f^\prime_{12}\rangle+\langle f^\prime_{21}\rangle)  -  \tfrac{\tau}{2}\langle c f_{12}+ c f_{21}\rangle  
     \\
     f_{\ell,\tilde{\ell}} &= -2 \sigma(\Theta_\ell)  \sigma^\prime (\Theta_{\tilde{\ell}} )
     \\
     f^\prime_{\ell,\tilde{\ell}} &= \big(\tfrac{\partial}{\partial \Theta_{1}} + \tfrac{\partial}{\partial \Theta_{2}} \big) f_{\ell,\tilde{\ell}} 
     \\
     \Theta_{\ell} &= c q + e \sqrt{1-q^2}  + \vct{\gamma}^{(\ell)}, \; \ell=1,2.
    \end{aligned}
\end{equation}

The relevant expectations of $\Theta$ of drift term are
\begin{equation*}
    \begin{aligned}
        \langle cf_{12}\rangle &= -4\langle c \Theta_1^2 \Theta_2\rangle=-4[q^3(m_4-3)+3q + \langle (\gamma^{(1)})^2 \rangle q + 2\langle \gamma^{(1)}\gamma^{(2)} \rangle q]
        \\
        \langle cf_{21}\rangle &= -4\langle c \Theta_1^2 \Theta_2\rangle=-4[q^3(m_4-3)+3q + \langle (\gamma^{(2)})^2 \rangle q + 2\langle \gamma^{(1)}\gamma^{(2)} \rangle q]
        \\
        \langle f'_{12}\rangle &= -4\langle 2\Theta_1\Theta_2 + \Theta_1^2 \rangle
        = -4[3+2\langle \gamma^{(1)} \gamma^{(2)}\rangle  + \langle (\gamma^{(1)})^2 \rangle]
        \\
        \langle f'_{21}\rangle &= -4\langle 2\Theta_1\Theta_2 + \Theta_1^2 \rangle
        = -4[3+2\langle \gamma^{(1)} \gamma^{(2)}\rangle  + \langle (\gamma^{(2)})^2 \rangle].
    \end{aligned}
    \end{equation*}

Thus the expression for $g$ is
    \begin{equation}
        g=-4 q^3(m_4-3),
    \end{equation}
where the impact of additive noise is reduced by the subtraction of $\langle cf \rangle$ and $\langle f' \rangle$. Next we deal with the diffusion term. The relevant expectations of $\Theta$ are
    \begin{equation*}
        \begin{aligned}
        \langle f_{12}^2 \rangle&= 16\langle \Theta_1^4 \Theta_2^4 \rangle
        \\
        &= 16[\langle \Theta^6 \rangle + (6\langle (\gamma^{(1)})^2 \rangle+8\langle \gamma^{(1)}\gamma^{(2)} \rangle+\langle (\gamma^{(2)})^2 \rangle)\langle \Theta^4 \rangle + (\langle (\gamma^{(1)})^4 \rangle+8\langle (\gamma^{(1)})^3 \gamma^{(2)} \rangle
        \\
        &+6\langle (\gamma^{(1)})^2 (\gamma^{(2)})^2 \rangle)\langle \Theta^2 \rangle + \langle (\gamma^{(1)})^4 (\gamma^{(2)})^2\rangle]
        \\
        &= 16[q^6(m_6-15) + 15q^4(1-q^2)(m_4-3) + 15 + (6\langle (\gamma^{(1)})^2 \rangle+8\langle \gamma^{(1)}\gamma^{(2)} \rangle+\langle (\gamma^{(2)})^2 \rangle) (q^4(m_4-3) + 3) 
        \\
        &+ \langle (\gamma^{(1)})^4 \rangle+8\langle (\gamma^{(1)})^3 \gamma^{(2)} \rangle+6\langle (\gamma^{(1)})^2 (\gamma^{(2)})^2 \rangle + \langle (\gamma^{(1)})^4 (\gamma^{(2)})^2\rangle ] 
        \\
        \langle f_{21}^2 \rangle
        &= 16[q^6(m_6-15) + 15q^4(1-q^2)(m_4-3) + 15 + (6\langle (\gamma^{(2)})^2 \rangle+8\langle \gamma^{(1)}\gamma^{(2)} \rangle+\langle (\gamma^{(1)})^2 \rangle) (q^4(m_4-3) + 3) 
        \\
        &+ \langle (\gamma^{(2)})^4 \rangle+8\langle (\gamma^{(2)})^3 \gamma^{(1)} \rangle+6\langle (\gamma^{(1)})^2 (\gamma^{(2)})^2 \rangle + \langle (\gamma^{(2)})^4 (\gamma^{(1)})^2\rangle ] 
        \\
        \langle f_{12}f_{21} \rangle &= 16 \langle \Theta_1^3 \Theta_2^3\rangle
        \\
        &=16[\langle \Theta^6 \rangle + (3\langle (\gamma^{(1)})^2 \rangle+9\langle \gamma^{(1)}\gamma^{(2)} \rangle+3\langle (\gamma^{(2)})^2 \rangle)\langle \Theta^4 \rangle + (3\langle (\gamma^{(1)})^3 \gamma^{(2)} \rangle
        \\
        &+9\langle (\gamma^{(1)})^2 (\gamma^{(2)})^2 \rangle+3\langle \gamma^{(1)} (\gamma^{(2)})^3 \rangle)\langle \Theta^2 \rangle + \langle (\gamma^{(1)})^3 (\gamma^{(2)})^3\rangle]
        \\
        &= 16[q^6(m_6-15) + 15q^4(1-q^2)(m_4-3) + 15 +  (3\langle (\gamma^{(1)})^2 \rangle+9\langle \gamma^{(1)}\gamma^{(2)} \rangle+3\langle (\gamma^{(2)})^2 \rangle) (q^4(m_4-3) + 3) 
        \\
        &+3\langle (\gamma^{(1)})^3 \gamma^{(2)} \rangle+9\langle (\gamma^{(1)})^2 (\gamma^{(2)})^2 \rangle+3\langle \gamma^{(1)} (\gamma^{(2)})^3 \rangle + \langle (\gamma^{(1)})^3 (\gamma^{(2)})^3\rangle ] .
    \end{aligned}
\end{equation*}

\begin{proof}[Proof of Claim 3]
  The expectation of $\langle (\gamma^{(1)})^{n_1} (\gamma^{(2)})^{n_2} \rangle$ is
\begin{equation*}
\begin{aligned}
 \langle (\gamma^{(1)})^2 \rangle &= \langle (\gamma^{(2)})^2 \rangle = \eta,\qquad \qquad\langle \gamma^{(1)} \gamma^{(2)} \rangle  =0,
 \\
 \langle (\gamma^{(1)})^4 \rangle &= \langle (\gamma^{(2)})^4 \rangle = 3\eta^2, \quad \qquad \langle (\gamma^{(1)})^2(\gamma^{(2)})^2 \rangle = \eta^2, \quad \langle (\gamma^{(1)})^3\gamma^{(2)} \rangle = \langle \gamma^{(1)}(\gamma^{(2)})^3 \rangle = 0,
 \\
 \langle (\gamma^{(1)})^4(\gamma^{(2)})^2 \rangle &= \langle (\gamma^{(1)})^2(\gamma^{(2)})^4 \rangle = 3\eta^3, 
\langle (\gamma^{(1)})^3(\gamma^{(2)})^3 \rangle = 0. 
\end{aligned}
    \end{equation*}

The diffusion term in (\ref{appendix-eq:45}) can be simplified as

\begin{equation}
    \Lambda = \tfrac{8\tau^2}{m}[(2+\eta)(q^6(m_6-15) + 15q^4(1-q^2)(m_4-3))  
    + 
    ( 7\eta^2 + 14\eta)(q^4(m_4-3) + 3)
    +3\eta^4 + 12\eta^3 + 18\eta^2 
    ].
\end{equation}

According to Theorem 2, the ODE for $q$ is
\begin{equation}
\begin{aligned}
    \tfrac{\text{d}q}{\text{d}t} &= q(qg-\tfrac{1}{2}\Lambda) -g
    \\
    &=
    4\tau (1-q^2)q^3(m_4 - 3) -\tfrac{4\tau^2}{m}q[(2+\eta)(q^6(m_6-15) + 15q^4(1-q^2)(m_4-3))  
    \\
    &\quad + 
    ( 7\eta^2 + 14\eta)(q^4(m_4-3) + 3)
    +3\eta^4 + 12\eta^3 + 18\eta^2 
    ].
    \\
\end{aligned}
\end{equation}

Denote $Q=q^2$, we have the ODE for $Q$

\begin{equation}
\label{appendix-eq:43}
    \begin{aligned}
    \tfrac{\text{d}Q}{\text{d}t} 
    &=
    4\tau (1-Q)Q^2(m_4 - 3) -\tfrac{4\tau^2}{m}Q[(2+\eta)(Q^3(m_6-15) + 15Q^2(1-Q)(m_4-3))  
    \\
    &\quad + 
    ( 7\eta^2 + 14\eta)(Q^2(m_4-3) + 3)
    +3\eta^4 + 12\eta^3 + 18\eta^2 
    ].
    \\
\end{aligned}
\end{equation}

Finally subtracting (\ref{appendix-eq:43}) at $\eta>0$ and $\eta=0$, we get (\ref{appendix-eq:46}) in Claim 3.

\end{proof}

 \begin{proof}[Proof of Claim 4]The expectation of $\langle (\gamma^{(1)})^{n_1} (\gamma^{(2)})^{n_2} \rangle$ is
\begin{equation*}
\begin{aligned}
 \langle (\gamma^{(1)})^2 \rangle &= \langle (\gamma^{(2)})^2 \rangle = \eta,\qquad \quad \qquad\langle \gamma^{(1)} \gamma^{(2)} \rangle  =-\eta,
 \\
 \langle (\gamma^{(1)})^4 \rangle &= \langle (\gamma^{(2)})^4 \rangle = 3\eta^2,  \qquad \quad \quad \langle (\gamma^{(1)})^2(\gamma^{(2)})^2 \rangle = \eta^2, \quad \langle (\gamma^{(1)})^3\gamma^{(2)} \rangle = \langle \gamma^{(1)}(\gamma^{(2)})^3 \rangle = -3\eta^2,
 \\
 \langle (\gamma^{(1)})^4(\gamma^{(2)})^2 \rangle &= \langle (\gamma^{(1)})^2(\gamma^{(2)})^4 \rangle = 15\eta^3, \; \; 
\langle (\gamma^{(1)})^3(\gamma^{(2)})^3 \rangle = -15\eta^3. 
\end{aligned}
    \end{equation*}

The diffusion term in (\ref{appendix-eq:45}) can be simplified as
\begin{equation}
    \Lambda = \tfrac{8\tau^2}{m}[(2+\eta)(q^6(m_6-15) + 15q^4(1-q^2)(m_4-3))  
    + 
    ( 8\eta^2 - 16\eta)(q^4(m_4-3) + 3)
    +30\eta^4 - 12\eta^3 + 6\eta^2 
    ].
\end{equation}

According to Theorem 2, the ODE for $q$ is
\begin{equation}
\begin{aligned}
    \tfrac{\text{d}q}{\text{d}t} &= q(qg-\tfrac{1}{2}\Lambda) -g
    \\
    &=
    4\tau (1-q^2)q^3(m_4 - 3) -\tfrac{4\tau^2}{m}q[2(q^6(m_6-15) + 15q^4(1-q^2)(m_4-3))  
    \\
    &\quad + 
    ( 8\eta^2 - 16\eta)(q^4(m_4-3) + 3)
    +30\eta^4 - 12\eta^3 + 6\eta^2 
    ].
    \\
\end{aligned}
\end{equation}

Denote $Q=q^2$, we have the ODE for $Q$

\begin{equation}
\label{appendix-eq:47}
    \begin{aligned}
    \tfrac{\text{d}Q}{\text{d}t} 
    &=
    4\tau (1-Q)Q^2(m_4 - 3) -\tfrac{4\tau^2}{m}Q[2(Q^3(m_6-15) + 15Q^2(1-Q)(m_4-3))  
    \\
    &\quad + 
    ( 8\eta^2 - 16\eta)(Q^2(m_4-3) + 3)
    +30\eta^4 - 12\eta^3 + 6\eta^2 
    ].
    \\
\end{aligned}
\end{equation}

    Similarly, we can calculate the subtraction of (\ref{appendix-eq:47}) for $\eta>0$ and $\eta=0$
    
    \begin{equation}
    \tfrac{\text{d}Q}{\text{d}t} \bigg|_{\eta>0}- \tfrac{\text{d}Q}{\text{d}t} \bigg|_{\eta=0} 
    = 
    -\tfrac{4\tau^2}{m}[( 8\eta^2 - 16\eta)(Q^2(m_4-3) + 3)
    +30\eta^4 - 12\eta^3 + 6\eta^2 
    ].
\end{equation}

When $\eta$ is relatively small i.e. $\eta \ll 1$, we can omit higher order terms of $\eta$ and get (\ref{appendix-eq:50}).

\end{proof}

\section{Supplementary experiments using ReLU as activation}

We  show additional experimental results on ReLu activation function besides the theoretical analysis of the case with quadratic activation function stated above.

\begin{figure}[htbp] 

\centering
\subfloat[Gradient for different $m_2$ with zero noise]
  {
    \label{appendix-fig:subfig71}\includegraphics[width=0.45\columnwidth]{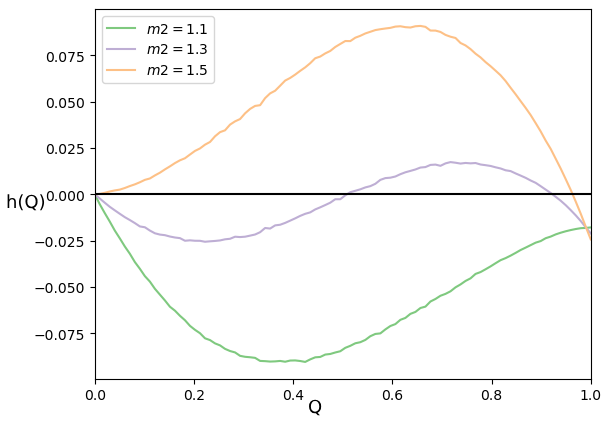}
  }
  \subfloat[Gradient for different strengths of  correlated additive noise]
  {
    \label{appendix-fig:subfig72}\includegraphics[width=0.45\columnwidth]{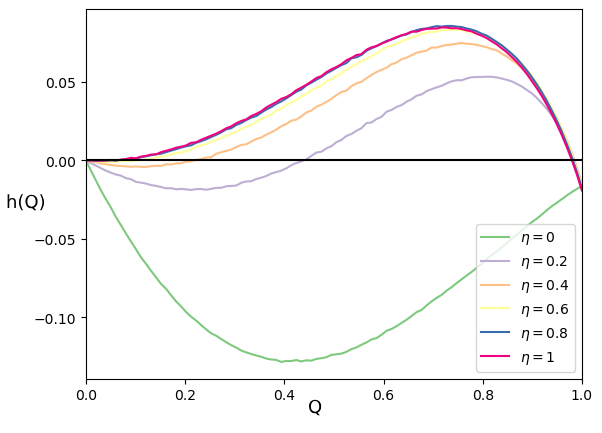}
  }\hfill
\caption{Gradient for different $m_2$ and different strengths of  correlated additive noise.}
\label{appendix-fig:7}
\end{figure}

Our observations in Fig. \ref{appendix-fig:2} and Fig. \ref{appendix-fig:6} can be verified experimentally when using $ReLU(x) = max(0, x)$ as activation. The results are plotted in Fig. \ref{appendix-fig:7} where $Q = q^2, h(Q) =\tfrac{\text{d}Q}{\text{d}t}/\tau $. 

As showed in Fig. \ref{appendix-fig:subfig71}, feature with higher second order moment is easier to retrieve. When $m_2=1.1$, there is only one fixed point $Q=0$. As $m_2$ increases, the fixed point near $Q=1$ emerges. When $m_2$ is larger than a threshold, the fixed point $Q=0$ become unstable. The observations are exactly the same as we mentioned before, the only difference is that the threshold for $\sigma(x) = ReLU(x)$ is larger than the threshold for $\sigma(x) = x^2$.

When $m_2$ is fixed, the effects of additive noise are considered and the gradient of different strengths of correlated additive noise is plotted in Fig. \ref{appendix-fig:subfig72}. As the strength of correlated additive noise increase, the fixed point near $Q=1$ emerges and the convergence region gradually expands, indicating that correlated additive noise can enhance the performance of feature recovery.


\begin{thebibliography}{26}
\providecommand{\url}[1]{#1}
\csname url@samestyle\endcsname
\providecommand{\newblock}{\relax}
\providecommand{\bibinfo}[2]{#2}
\providecommand{\BIBentrySTDinterwordspacing}{\spaceskip=0pt\relax}
\providecommand{\BIBentryALTinterwordstretchfactor}{4}
\providecommand{\BIBentryALTinterwordspacing}{\spaceskip=\fontdimen2\font plus
\BIBentryALTinterwordstretchfactor\fontdimen3\font minus
  \fontdimen4\font\relax}
\providecommand{\BIBforeignlanguage}[2]{{%
\expandafter\ifx\csname l@#1\endcsname\relax
\typeout{** WARNING: IEEEtran.bst: No hyphenation pattern has been}%
\typeout{** loaded for the language `#1'. Using the pattern for}%
\typeout{** the default language instead.}%
\else
\language=\csname l@#1\endcsname
\fi
#2}}
\providecommand{\BIBdecl}{\relax}
\BIBdecl

\bibitem{devlin_bert_2019}
J.~D. M.-W.~C. Kenton and L.~K. Toutanova, ``Bert: Pre-training of deep
  bidirectional transformers for language understanding,'' in \emph{NAACL},
  2019, pp. 4171--4186.

\bibitem{chen_simple_2020}
T.~Chen, S.~Kornblith, M.~Norouzi, and G.~Hinton, ``A simple framework for
  contrastive learning of visual representations,'' in \emph{ICML}, 2020, pp.
  1597--1607.

\bibitem{he_momentum_2020}
K.~He, H.~Fan, Y.~Wu, S.~Xie, and R.~Girshick, ``Momentum contrast for
  unsupervised visual representation learning,'' in \emph{CVPR}, 2020.

\bibitem{grill_bootstrap_2020}
J.-B. Grill, F.~Strub, F.~Altché, C.~Tallec, P.~Richemond, E.~Buchatskaya,
  C.~Doersch, B.~Avila~Pires, Z.~Guo, M.~Gheshlaghi~Azar, B.~Piot,
  k.~kavukcuoglu, R.~Munos, and M.~Valko, ``Bootstrap your own latent - a new
  approach to self-supervised learning,'' in \emph{NeurIPS}, vol.~33, 2020, pp.
  21\,271--21\,284.

\bibitem{zbontar_barlow_2021}
J.~Zbontar, L.~Jing, I.~Misra, Y.~{LeCun}, and S.~Deny, ``Barlow twins:
  Self-supervised learning via redundancy reduction,'' in \emph{ICML}, 2021,
  pp. 12\,310--12\,320.

\bibitem{bardes_vicreg_2022}
A.~Bardes, J.~Ponce, and Y.~Lecun, ``{VICReg}: Variance-invariance-covariance
  regularization for self-supervised learning,'' in \emph{ICLR}, 2022.

  \bibitem{saunshi_theoretical_2019}
N.~Saunshi, O.~Plevrakis, S.~Arora, M.~Khodak, and H.~Khandeparkar, ``A
  theoretical analysis of contrastive unsupervised representation learning,''
  in \emph{ICLR}, vol.~97, 2019, pp. 5628--5637.

  \bibitem{tao_exploring_2022}
C.~Tao, H.~Wang, X.~Zhu, J.~Dong, S.~Song, G.~Huang, and J.~Dai, ``Exploring
  the equivalence of siamese self-supervised learning via a unified gradient
  framework,'' in \emph{CVPR}, 2022, pp. 14\,431--14\,440.

  \bibitem{haochen_provable_2021}
J.~Z. {HaoChen}, C.~Wei, A.~Gaidon, and T.~Ma, ``Provable guarantees for
  self-supervised deep learning with spectral contrastive loss,'' in
  \emph{NeurIPS}, vol.~34, 2021, pp. 5000--5011.

  \bibitem{huang_towards_2023}
W.~Huang, M.~Yi, X.~Zhao, and Z.~Jiang, ``Towards the generalization of
  contrastive self-supervised learning,'' in \emph{ICLR}, 2023.

  \bibitem{tian_understanding_2022}
Y.~Tian, ``Understanding the role of nonlinearity in training dynamics of
  contrastive learning,'' \emph{arXiv preprint arXiv:2206.01342}, 2022.

 \bibitem{jing_understanding_2022}
L.~Jing, P.~Vincent, Y.~{LeCun}, and Y.~Tian, ``Understanding dimensional
  collapse in contrastive self-supervised learning,'' in \emph{ICLR}, 2022.

   \bibitem{tian_understanding_2022-1}
Y.~Tian, ``Understanding deep contrastive learning via coordinate-wise
  optimization,'' in \emph{NeurIPS}, 2022.

  \bibitem{wang2018subspace}
C.~Wang, Y.~C. Eldar, and Y.~M. Lu, ``Subspace estimation from incomplete
  observations: A high-dimensional analysis,'' \emph{IEEE Journal of Selected
  Topics in Signal Processing}, vol.~12, no.~6, pp. 1240--1252, 2018.

\bibitem{wang2017scaling}
C.~Wang and Y.~Lu, ``The scaling limit of high-dimensional online independent
  component analysis,'' in \emph{NeurIPS}, 2017.

\bibitem{goldt_dynamics_2019}
S.~Goldt, M.~Advani, A.~M. Saxe, F.~Krzakala, and L.~Zdeborová, ``Dynamics of
  stochastic gradient descent for two-layer neural networks in the
  teacher-student setup,'' in \emph{NeurIPS}, vol.~32, 2019.

\bibitem{mei_mean_2018}
S.~Mei, A.~Montanari, and P.-M. Nguyen, ``A mean field view of the landscape of
  two-layer neural networks,'' \emph{PNAS}, vol. 115, no.~33, pp. E7665--E7671,
  2018.

\bibitem{wang_solvable_2019}
C.~Wang, H.~Hu, and Y.~Lu, ``A solvable high-dimensional model of {GAN},'' in
  \emph{NeurIPS}, vol.~32, 2019.

  \bibitem{khemakhem_variational_2020}
I.~Khemakhem, D.~Kingma, R.~Monti, and A.~Hyvarinen, ``Variational autoencoders
  and nonlinear ica: A unifying framework,'' in \emph{AISTATS}, 2020, pp.
  2207--2217.

\bibitem{saad_exact_1995}
D.~Saad and S.~A. Solla, ``Exact solution for on-line learning in multilayer
  neural networks,'' vol.~74, no.~21, p. 4337, 1995.

\bibitem{keller_stochastic_1973}
J.~Keller and H.~McKean, \emph{Stochastic Differential Equations}, ser.
  SIAM-AMS proceedings.\hskip 1em plus 0.5em minus 0.4em\relax AMS, 1973.

\bibitem{wang2017scaling}
C.~Wang and Y.~Lu, ``The scaling limit of high-dimensional online independent
  component analysis,'' in \emph{NeurIPS}, 2017.

\bibitem{veiga_phase_2022}
R.~Veiga, L.~Stephan, B.~Loureiro, F.~Krzakala, and L.~Zdeborová, ``Phase
  diagram of stochastic gradient descent in high-dimensional two-layer neural
  networks,'' in \emph{NeurIPS}, vol.~35, 2022, pp. 23\,244--23\,255.

\bibitem{wen_toward_2021}
Z.~Wen and Y.~Li, ``Toward understanding the feature learning process of
  self-supervised contrastive learning,'' in \emph{ICML}, vol. 139, 2021, pp.
  11\,112--11\,122.

\bibitem{yang_understanding_2023}
R.~Yang, X.~Li, B.~Jiang, and S.~Li, ``Understanding representation
  learnability of nonlinear self-supervised learning,'' in \emph{AAAI},
  vol.~37, 2023, pp. 10\,807--10\,815.

\bibitem{chen_exploring_2021}
X.~Chen and K.~He, ``Exploring simple siamese representation learning,'' in
  \emph{CVPR}, 2021, pp. 15\,750--15\,758.

\end{thebibliography}
\end{document}